\documentclass[final,5p,times,twocolumn]{elsarticle}

\usepackage{amssymb}
\usepackage{amsmath}
\usepackage{booktabs}
\usepackage{tabularx}
\usepackage{subcaption}
\usepackage{makecell}
\usepackage{multirow}
\usepackage{bm}
\usepackage{listings}
\usepackage{enumitem}
\usepackage{placeins}
\usepackage[svgnames]{xcolor}
\usepackage[hidelinks]{hyperref}

\definecolor{colorIncluded}{RGB}{188, 221, 149}
\definecolor{colorDiscarded}{RGB}{230, 160, 155}
\definecolor{linkColor}{RGB}{14, 116, 179}
\definecolor{highlightChangeColor}{RGB}{50, 86, 168}
\newcommand{\highlight}[1]{{\color{highlightChangeColor}{#1}}}
\newcommand{\legendLetterLarge}[1]{\raisebox{-1pt}{\includegraphics[height=8pt]{img/#1.pdf}}}
\newcommand{\legendLetterSmall}[1]{\raisebox{-1pt}{\includegraphics[height=7pt]{img/#1.pdf}}}

\journal{Computers \& Graphics}

\begin{document}

\begin{frontmatter}

\title{Leveraging LLMs for Semi-Automatic Corpus Filtration in Systematic Literature Reviews}

\author[ukon]{Lucas Joos\corref{cor1}}
\ead{lucas.joos@uni-konstanz.de}
\author[ukon]{Daniel A. Keim}
\ead{keim@uni-konstanz.de}
\author[ukon]{Maximilian T. Fischer}
\ead{max.fischer@uni-konstanz.de}

\affiliation[ukon]{
            organization={University of Konstanz},
            city={Konstanz},
            country={Germany}}

\cortext[cor1]{Corresponding author}

\begin{abstract}
The creation of systematic literature reviews (SLR) is critical for analyzing the landscape of a research field and guiding future research directions.
However, retrieving and filtering the literature corpus for an SLR is highly time-consuming and requires extensive manual effort, as keyword-based searches in digital libraries often return numerous irrelevant publications.
In this work, we propose a pipeline leveraging multiple large language models (LLMs), classifying papers based on descriptive prompts and deciding jointly using a consensus scheme.
The entire process is human-supervised and interactively controlled via our open-source visual analytics web interface, LLMSurver, which enables real-time inspection and modification of model outputs.
We evaluate our approach using ground-truth data from a recent SLR comprising over 8,000 candidate papers, benchmarking both open and commercial state-of-the-art LLMs from mid-2024 and fall 2025.
Results demonstrate that our pipeline significantly reduces manual effort while achieving lower error rates than single human annotators.
Furthermore, modern open-source models prove sufficient for this task, making the method accessible and cost-effective.
Overall, our work demonstrates how responsible human-AI collaboration can accelerate and enhance systematic literature reviews within academic workflows.
\end{abstract}

\begin{keyword}

Systematic Literature Review
\sep
Large Language Models
\sep
Corpus Filtration
\sep
Automatic Literature Filtering
\sep
Human-AI Collaboration
\sep
Research Automation

\end{keyword}

\end{frontmatter}

\section{Introduction}
\label{sec:introduction}

\begin{figure*}[!ht]
    \includegraphics[width=1\linewidth, alt={schematic}]{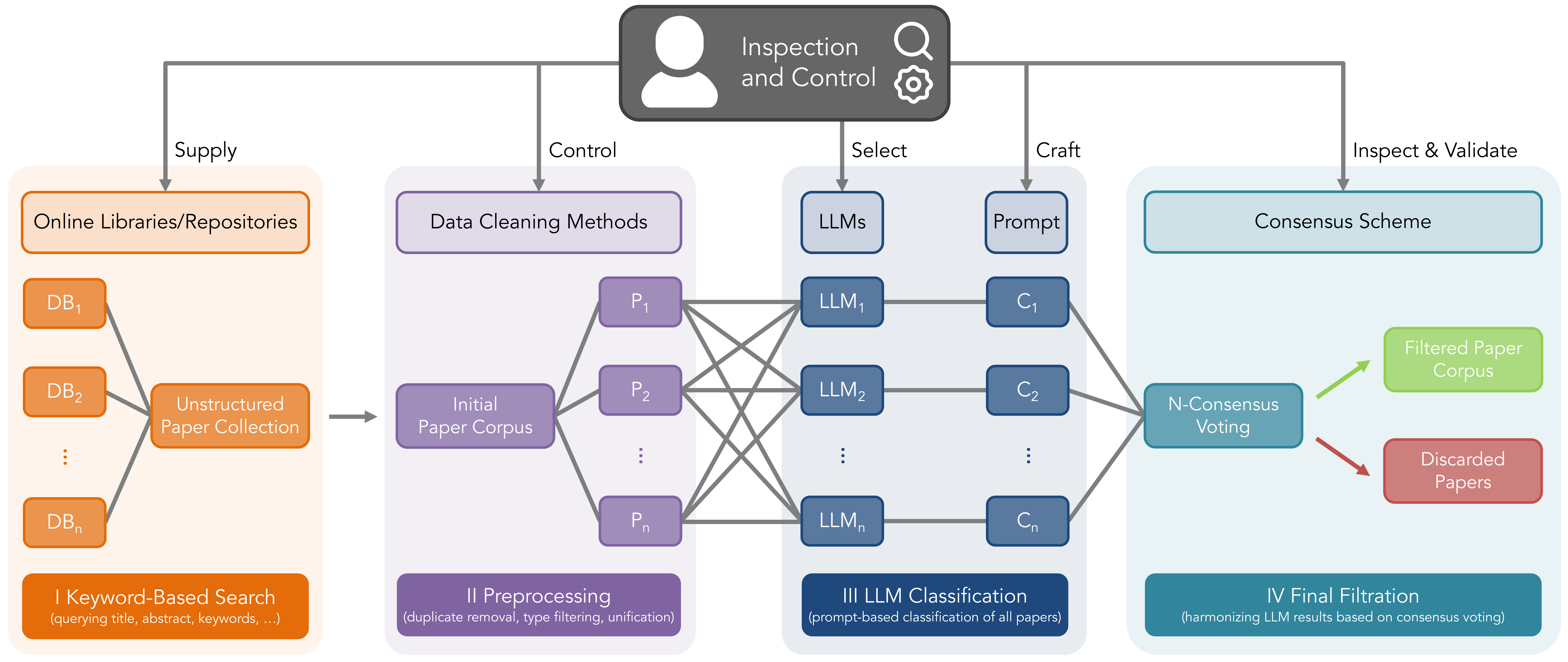}
    \centering
    \caption{Schematic overview of leveraging LLM-based agents for structured literature filtration in a systematic literature review. Keyword-based search in online libraries generates a large set of candidate papers that are preprocessed, and classified by multiple LLMs based on title and abstract using a customized prompt. A consensus voting scheme determines the final inclusion or rejection. The user keeps the control through inspecting and adapting all steps, including the initial database selection and search, preprocessing, LLM models, prompts, and the consensus scheme.
    }
   \label{fig:teaser}
\end{figure*}

Literature reviews, and especially \emph{systematic literature reviews} (SLRs), are often regarded as the \textit{gold standard} for conducting structured and comprehensive research syntheses in academia~\cite{egger2008systematic, Davis.SystematicReviews.2014}.
They offer a transparent and reproducible methodology for organizing and synthesizing existing research findings, providing a thorough and reliable overview of the landscape for a specific research area~\cite{Nightingale.SLRGuide.2009}.
The practice of conducting such reviews dates back at least to the 18th century~\cite{VanDinter.SLRAutomation.2021} and continues to play a crucial role in identifying research gaps, outlining future directions, and ensuring consistency and reliability in academic inquiry~\cite{Lame.SLRIntroduction.2019}.
Despite their importance, the creation of SLRs remains a highly manual and time-consuming endeavor~\cite{SilvaJunior.RoadmapSLR.2021}.
Researchers must possess a strong understanding of their field to effectively search for, categorize, and code a large number of potentially relevant publications.
While this process may be manageable within narrowly defined research areas, it becomes increasingly challenging in broader or fast-evolving domains where publication volumes grow rapidly. For large-scale SLRs, the process can extend over several months.
Egger et al. outline the standard SLR process in eight stages: (1) defining the research question, (2) setting inclusion and exclusion criteria, (3) locating potentially relevant work, (4) selecting publications, (5) assessing their content, (6) extracting information, (7) presenting results, and (8) interpreting findings~\cite{egger2008systematic}.
A widely adopted framework, \textit{PRISMA}~\cite{prisma2009}, formalizes the steps of corpus retrieval (stages 3–6) through keyword-based search, duplicate removal, manual screening of titles and abstracts, and full-text review.
Among these stages, manual title and abstract screening is particularly labor-intensive, especially in research areas where keyword-based search returns a vast amount of ambiguous or irrelevant results.
Wallace et al.~\cite{Wallace.SemiAutomatedScreeningSystematicReview.2010} report that an experienced reviewer can manually screen approximately two papers per minute based on title and abstract.
At this pace, reviewing a corpus of roughly 8,000 potentially relevant papers for a large SLR requires around 66 person-hours, equivalent to about one and a half full workweeks of \emph{uninterrupted} effort.
In reality, factors such as fatigue, reduced focus, quality control through dual verification, and competing commitments significantly extend this duration, often resulting in several months of screening time before analysis can even begin.
Given the repetitive yet cognitively demanding nature of this task and the rapid increase in academic output, it is reasonable to ask whether and how this process can be optimized.

Efforts to automate parts of this classification process date back to the mid-2000s~\cite{VanDinter.SLRAutomation.2021}.
However, the emergence of LLMs has recently opened new opportunities for automating the initial filtration of literature during SLR creation.
These models offer several advantages: (1) they capture subtle semantic nuances, making it possible to reach sufficient levels of recall and precision, (2) they operate at higher speed and lower cost compared to manual human screening, and (3) they continue to improve rapidly in quality and availability, including open-access models.
Open LLMs, particularly smaller variants, can be deployed locally on institutional servers or personal computers, ensuring data privacy and eliminating dependency on external paid providers.
Consequently, such models hold strong potential to enhance both the efficiency and overall quality of academic work involved in conducting systematic literature reviews.

With our work, we aim to investigate how effectively LLMs can support researchers in creating SLRs, particularly in filtering an initial corpus retrieved through keyword-based online library searches.
To achieve this, we propose a human-AI collaboration pipeline that employs multiple LLMs to classify papers and reach a final decision through a consensus scheme.
Through a visual analytics application that implements the semi-automatic pipeline, the user supervises the process, refines it (for example, the prompts, involved LLMs, and consensus scheme), and evaluates the results.
We assess our corpus filtration pipeline using real-world data comprising more than 8,000 papers that were retrieved for a recent survey paper~\cite{joos2025visualnetworkanalysisimmersive} through keyword-based search and for which human ground-truth labeling is available.
To examine the evolution and performance changes of LLMs, we evaluated the pipeline on the given dataset using five LLMs that represented the state of the art in mid-2024 (as reported in our initial work~\cite{joos2025cutting}), and again with 13 modern models from fall 2025, eight of which are openly available.
A detailed analysis of the LLM results and multiple consensus schemes highlights their suitability for the task and shows that smaller open models also achieve strong results.
Additionally, for one of the open LLMs, we examine how the results vary with modified prompts.
In summary, we make the following contributions:
\begin{itemize}[leftmargin=1.2em]
\setlength\itemsep{-0.2em}
   \item A \textbf{conceptual pipeline} for the semi-automatic paper corpus filtration leveraging LLMs voting in a consensus scheme.
	
  \item The open-source \textbf{application} \href{https://github.com/dbvis-ukon/LLMSurver}{{\color{linkColor}LLMSurver}} (\href{https://llmsurver.dbvis.de}{{\color{linkColor}Web Demo}}), implementing the pipeline and making it accessible to others
	
  \item A thorough \textbf{evaluation} of the pipeline using previous (mid-2024) and modern (fall 2025) LLMs for a large SLR (8.3k papers) and an analysis exploring how different prompts influence the results..

  \item A detailed \textbf{discussion} exploring the current challenges, the emerging potentials, as well as the future trajectories of AI-driven literature filtration.
\end{itemize}

\section{Related Work}
\label{sec:related_work}

With the recent advances in machine learning, and particularly in large language models, an increasing number of studies have demonstrated how these models can support various stages of the scientific publishing process~\cite{Wallace.SemiAutomatedScreeningSystematicReview.2010, SilvaJunior.RoadmapSLR.2021, lund2023chatgpt}.
Examples of such tasks include generating peer reviews to improve one's own work~\cite{tyser2024aidriven}, reformulating text for clarity~\cite{gilat2023how}, identifying and mitigating biased arguments~\cite{huang2023role}, and detecting research gaps within a given domain~\cite{lund2023chatgpt}.

Beyond these general publishing-related applications, LLMs have recently been shown to assist in several stages of conducting literature reviews~\cite{Sami.SystematicLiteratureReviewAIAgents.2024}.
Although a substantial body of research describes established methodologies for performing literature reviews~\cite{Nightingale.SLRGuide.2009, Davis.SystematicReviews.2014, egger2008systematic, Lame.SLRIntroduction.2019}, many of these approaches remain highly repetitive and labor-intensive.
To address this challenge, recent studies have investigated how agent-based systems can support the formulation of research questions, corpus filtering, and domain-specific searches~\cite{Whitfield.ElicitAILiteratureReview.2023, Sami.SystematicLiteratureReviewAIAgents.2024, huang2023role}.
Other approaches explore the use of LLMs for keyword generation and retrieval through retrieval-augmented generation (RAG)~\cite{Agarwal.LitLLM.2024}, or, more broadly, how machine learning~\cite{Wallace.SemiAutomatedScreeningSystematicReview.2010, VanDinter.SLRAutomation.2021, SilvaJunior.RoadmapSLR.2021} and LLM-based methods~\cite{Antu.LLMLiteratureReview.2023, susnjak2023prisma, rathi2023p21, bolanos2024artificial, hawkins2024literature, peinl2024usingLLM} can enhance different steps of the overall review process.
Furthermore, LLMs have also been applied to support the summarization phase of literature reviews~\cite{li2024chatcite}.

A specific aspect that has received comparatively little attention is the accurate filtering and classification of a large corpus of potentially relevant research to accelerate the \textit{paper pre-selection process}.
This step is particularly important in research areas where keyword-based filtering proves difficult, for instance, due to semantic ambiguity or the frequent reuse of terms across contexts.
Haryanto~\cite{haryanto2024llassist} investigated the use of LLMs for this task, focusing on individual model predictions.
Directly using modern chat-based LLMs for paper selection in SLR has not yielded human-level results so far~\cite{Chen962}.
More recently, several automated tools for SLR generation based on LLMs have been proposed~\cite{scherbakov2024emergence, gana2024leveraging, jafari2024streamlining, susnjak2024automating}.
Among these, Gehrmann et al.~\cite{gehrmann2024large} introduced the only LLM-based automated pre-selection method, demonstrating that negative prompting can enhance classification accuracy.
Very recently, approaches that use single LLMs in an automated setting have been proposed~\cite{SunSharPei202530, Scherbakov2025}.

In contrast to previous work, we present a pipeline for classifying and filtering large paper corpora by leveraging \emph{multiple} LLMs and combining their outputs through a consensus scheme to derive final decisions.
Building upon our initial study~\cite{joos2025cutting}, in which we compared manual human classification with an LLM-based pipeline using only a few state-of-the-art models available in mid-2024, we now extend our investigation to 13 models, most of which are openly accessible, and compare their performance.
Furthermore, we examine how variations in prompts influence the final outcomes.
Our findings provide deeper insights into the capabilities of modern LLMs (both open and commercial), the design of effective consensus schemes, the role of prompt engineering, and the overall reliability and accuracy of the proposed approach.

\section{Methodology}
\label{sec:methodology}

\begin{figure}
    \fbox{
   \begin{minipage}[b]{0.95\linewidth}
   \setlength{\parskip}{2pt}
   \scriptsize \fontfamily{qag} \fontsize{6}{8}\selectfont
    You are a professor in computer science conducting a literature review.
    Please decide and classify if the following paper belongs to a specific research direction or not.
    For this, you are provided with the title and the abstract, which should give you sufficient information for an informed and accurate decision.
    
    The research direction is the topic of "TITLE".
    
    Therefore include papers that deal with ASPECT\_1, ASPECT\_2, ... Examples of ASPECT\_1 are: term 1, term 2, \ldots
    
    You MUST discard papers that EXCLUSION\_EXCEPTION\_1, \ldots
    
    You MUST include papers that INCLUSION\_EXCEPTION\_1, \ldots
    
    Below is the title and abstract. You must only answer with INCLUDE or DISCARD and a 2-sentence reason of why.
    \end{minipage}
    }
    \caption{An exemplary prompt template for individual LLM agents.}
    \label{fig:prompt_format}
\end{figure}

In our work, we aim to investigate how LLMs can accelerate the corpus retrieval process for SLR creation without compromising the quality of the final results.
To this end, we propose a pipeline (see \autoref{fig:teaser}) built upon the well-established PRISMA~\cite{prisma2009} framework, which we extend through close collaboration between AI agents (LLMs) and human researchers.
The initial steps of our proposed pipeline follow the classical paper retrieval process (see \autoref{sec:methodology}).
Multiple online libraries are searched for relevant publications using defined keywords and additional criteria such as venue and publication date.
This results in an initial paper corpus.
Subsequent automated filtering steps, including duplicate removal, data unification, and type-based filtering, further refine this corpus.
Typically, the remaining papers are then manually screened, with researchers reviewing titles and abstracts to exclude irrelevant works.
In our approach, however, this screening step is performed by LLMs, which individually classify papers using a customized prompt (as illustrated in \autoref{fig:prompt_format}), the paper title, and its abstract.

Relying solely on individual LLMs for final inclusion decisions without human oversight poses significant risks.
Therefore, human researchers remain an essential part of the process.
They iteratively design and refine prompts, inspect sampled LLM outputs through a visual-interactive interface, and evaluate results until a satisfactory level of quality is achieved.
Examining the reasoning provided by the LLMs for their classifications often offers valuable insights for prompt refinement and model evaluation.
Moreover, our approach leverages not only one but multiple LLMs with diverse architectures, training objectives, and configurations, which provide complementary strengths and perspectives.
Once a promising subset of models has been identified and the preliminary results meet quality expectations, each LLM classifies all papers in the corpus.

With $n$ LLMs involved, each paper receives $n$ potentially differing classifications.
These individual outputs are then harmonized through a consensus scheme to obtain the final decision.
Various strategies can be employed for consensus formation, such as majority voting or the use of an additional LLM that analyzes the individual decisions and rationales to derive a final decision.
In our case, minimizing the exclusion of relevant publications is of highest importance.
Hence, our consensus scheme includes a paper if any of the participating LLMs recommends its inclusion.
Only if all LLMs vote for rejection is the paper excluded.
To define an appropriate consensus scheme and select the contributing models, our pipeline integrates a visual-interactive process that assists researchers in analyzing LLM results, identifying similarities and discrepancies between models, and constructing a consensus approach aligned with their objectives.
This process is supported by our visual-interactive tool \textit{LLMSurver}, as described in \autoref{sec:application}.

\begin{figure*}[!th]
    \centering
    \vspace*{-6mm}
    \hspace*{-7mm}
    \includegraphics[width=1.07\linewidth]{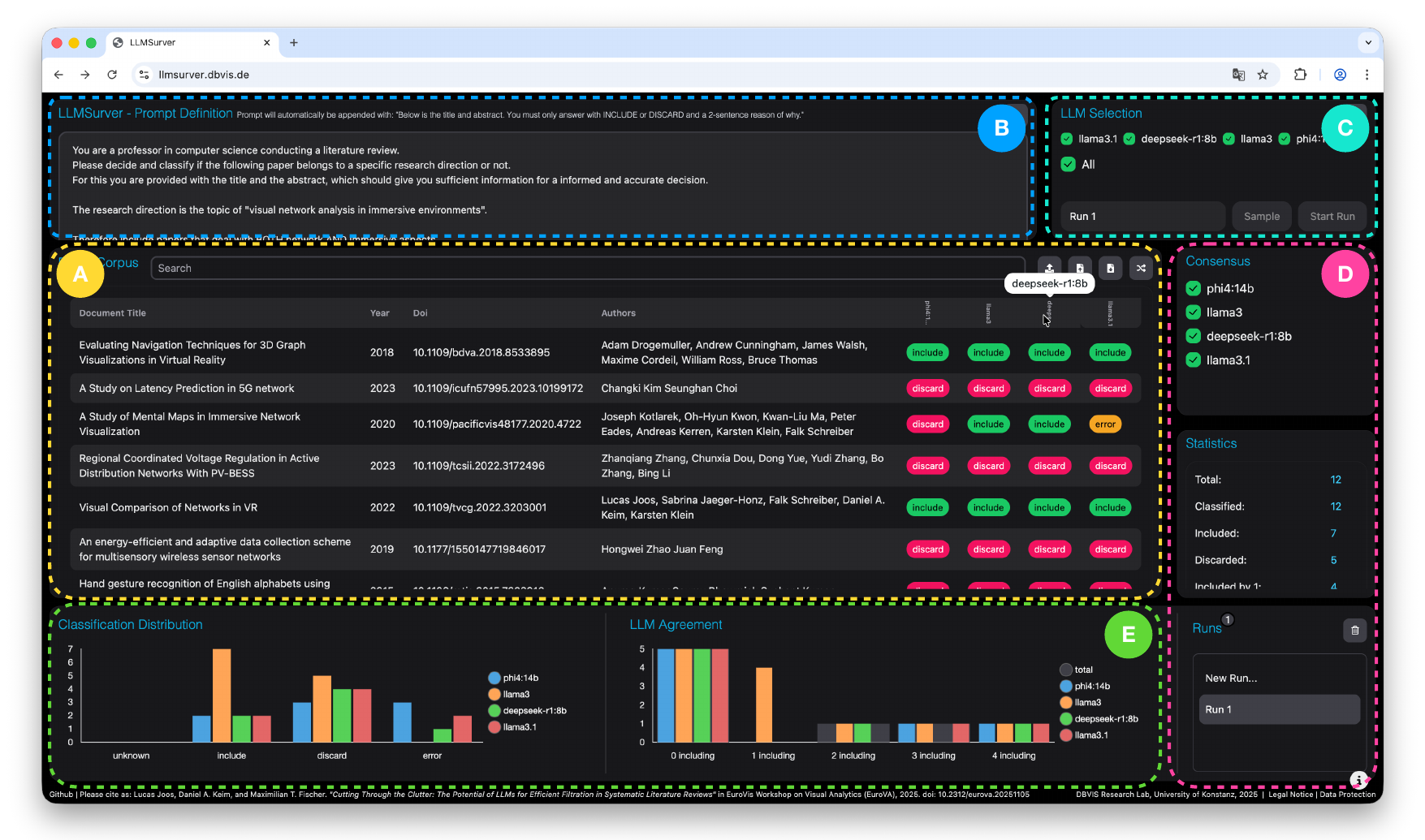}
    \vspace*{-6mm}
    \caption{Overview of the user interface of our open-source application \textbf{LLMSurver}, which implements the proposed pipeline. The interface includes a paper table \legendLetterSmall{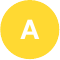}, a prompt editor \legendLetterSmall{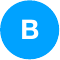}, controls for LLM selection and classification execution \legendLetterSmall{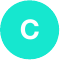}, a consensus scheme and statistics view \legendLetterSmall{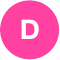}, and visual plots for comparative analysis \legendLetterSmall{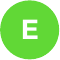}. The application can be used freely at \href{https://llmsurver.dbvis.de}{{\color{linkColor}\texttt{https://llmsurver.dbvis.de}}}.}
    \label{fig:application}
\end{figure*}

Ultimately, the combination of LLM-based classification, consensus harmonization, and continuous human involvement results in a filtered paper corpus suitable for further analysis and SLR development.
Although false exclusions may still occur despite the conservative consensus strategy and multi-model setup, similar errors can also arise in manual screening.
Such papers can typically be recovered in a subsequent snowballing step~\cite{wohlin2014guidelines}.
Therefore, a small number of false removals during the LLM-based filtering stage may be considered acceptable.
Furthermore, the corpus will most likely still contain papers that are not relevant to the study.
However, the initial filtering step substantially reduces the number of candidates that need to be manually reviewed by the researchers, which considerably speeds up the overall process.

\section{LLMSurver}
\label{sec:application}

We developed the interactive open-source web-application \textbf{LLMSurver}, which implements the proposed pipeline.
This application demonstrates the practical feasibility of our approach and provides direct support for researchers conducting their own literature surveys.
The tool allows users to automatically perform classification with multiple LLMs in parallel for a given dataset and prompt.
In addition, the user interface (UI) enables users to analyze results, refine prompts, compare consensus schemes, and export the outcomes.

\begin{figure*}[!th]
    \centering
    \begin{subfigure}[b]{0.49\textwidth}
        \centering
        \includegraphics[width=\textwidth]{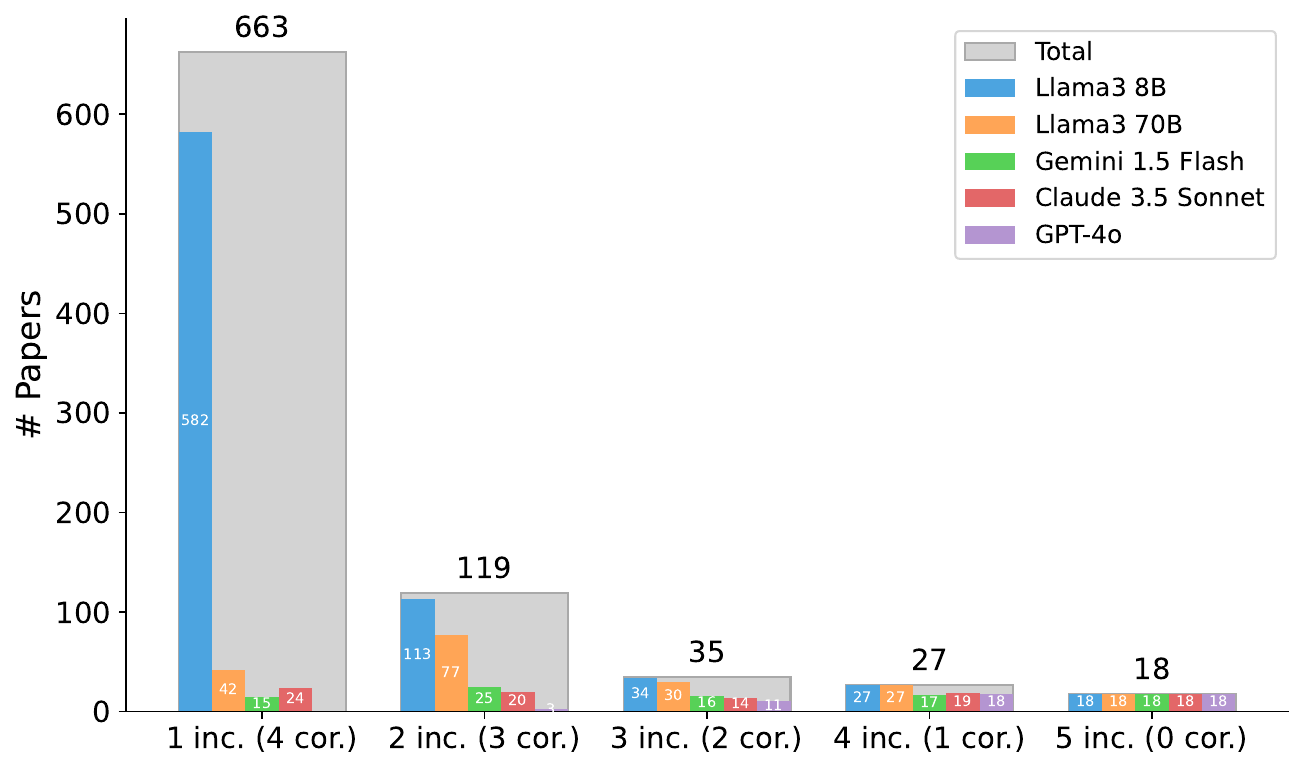}
    \end{subfigure}
    \hfill
    \begin{subfigure}[b]{0.49\textwidth}
        \centering
        \includegraphics[width=\textwidth]{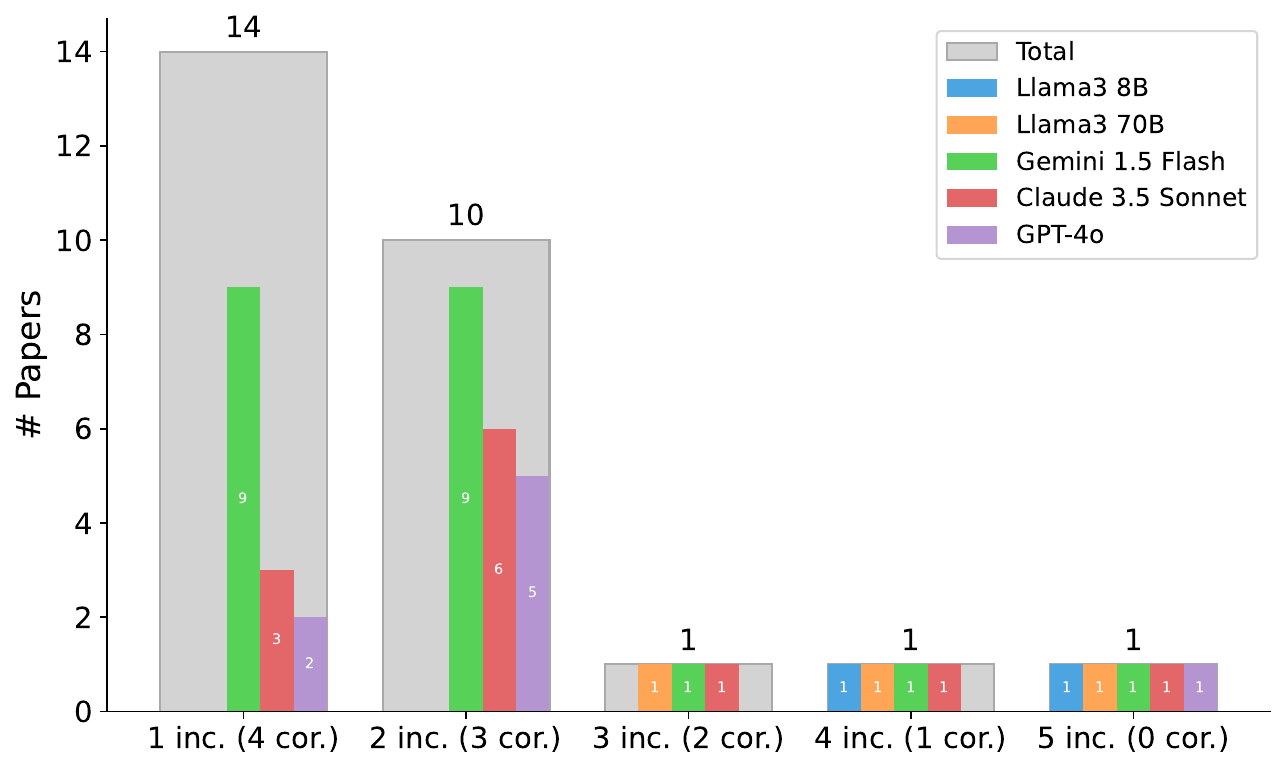}
    \end{subfigure}
    \caption{Number of papers (gray background) that were \emph{incorrectly (inc.)} classified as \textbf{included} (left) or \textbf{excluded} (right) by the \textbf{Mid-2024} LLM agents, grouped by the number of agents involved in each incorrect decision. The individual bars indicate how often a specific agent contributed to a wrong decision. On the far right, only one paper is misclassified by all agents (and therefore lost permanently), illustrating that N-Consensus voting is advantageous when prioritizing recall.}
    \label{fig:llm-inclusion-exclusion}
\end{figure*}

The web application is fully containerized and features a single-page \texttt{React} frontend that is easy to deploy, maintain, and extend.
The system performs all operations locally within the browser environment.
LLM API calls are the only exception and can utilize local or custom inference deployments with authentication tokens.
This design eliminates the need for a backend and guarantees that no data is transmitted to external servers.
The UI follows a dashboard layout with visually distinct components that correspond to the steps of the pipeline (see~\autoref{fig:application}).
The main table \legendLetterLarge{a} displays paper details from the corpus, populated by uploading \texttt{BibTeX} files, code, or \texttt{DOI} numbers.
In the latter case, the publication information is retrieved from online resources.
A prompt editor component \legendLetterLarge{b} enables users to create and refine classification prompts for selected LLMs in component \legendLetterLarge{c}.
Users can register new LLMs, either local or remote, by entering the required information, such as API keys or hostnames.
Additional parameters, such as the temperature, can be added as well.
Classifications can be executed on selected subsets (samples) for testing or on the entire corpus, with intermediate results stored for later inspection.
Classification outputs are visualized in the main table and can be exported as a \texttt{CSV} file.
Users can also inspect individual LLM outputs and reasonings, which is particularly useful in cases of ambiguous results indicated by an orange \texttt{error} icon.
The consensus component \legendLetterLarge{d} supports selecting a specific run, displays additional statistics, and enables LLM selection for consensus building, with the corresponding results displayed in the main table.
To facilitate decision-making, component \legendLetterLarge{e} provides two charts.
The first visualization shows the distribution of classifications across LLMs, while the second visualizes agreement levels and highlights outliers that may lower the quality of the consensus (similar to \autoref{fig:llm-inclusion-exclusion} and \autoref{fig:new-llm-inclusion-exlusion}).

\begin{figure*}[!ht]
    \centering
    \includegraphics[width=1\linewidth]{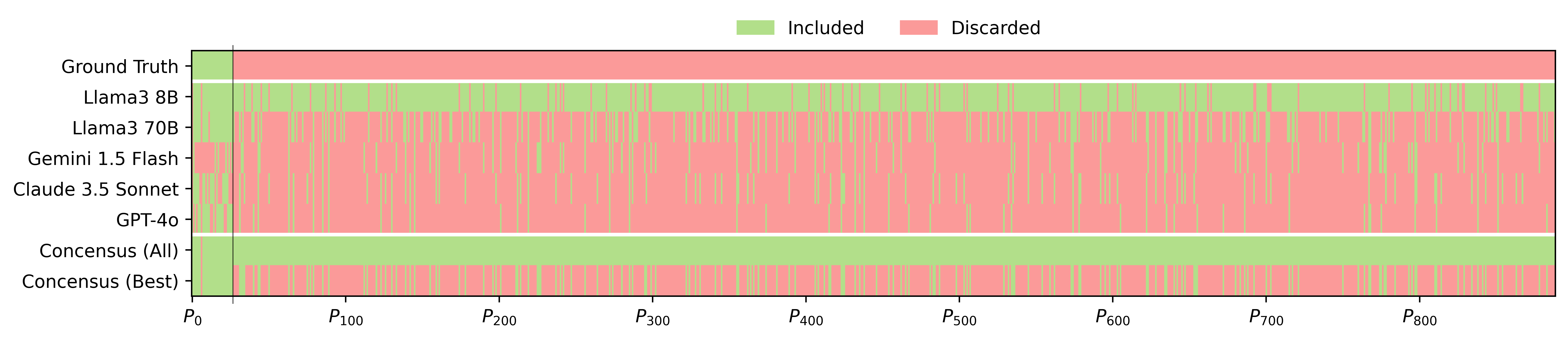}
    \caption{Overview of pairwise comparisons showing \emph{incorrect} decisions by the agents for the \textbf{Mid-2024} models. The ground-truth classification of each paper is shown on the top row (left side: included, right side: discarded), followed by the decisions of individual agents, and finally by the two consensus methods (all models and top-3 models). Incorrect exclusions (FN) appear as \textcolor{colorDiscarded}{red discarded} lines on the left, whereas incorrect inclusions (FP) appear as \textcolor{colorIncluded}{green included} lines on the larger right side. Note the single incorrectly discarded paper (FN) for the consensus methods at the bottom left, which reduces recall.}
    \label{fig:matrix}
\end{figure*}

The tool is highly adaptable and can be used for various other purposes.
It supports the integration of custom consensus strategies and additional decision-making visualizations.
We provide the source code openly under \href{https://github.com/dbvis-ukon/LLMSurver}{{\color{linkColor}\texttt{https://github.com/dbvis-ukon/LLMSurver}}} (MIT License).
A freely available deployed version of the tool can be accessed at \href{https://llmsurver.dbvis.de}{{\color{linkColor}\texttt{https://llmsurver.dbvis.de}}}, which allows convenient, privacy-preserving evaluation of our approach without the need for local deployment.
A short tutorial that appears upon the first start of the application facilitates onboarding and helps users to fully explore the available functionality with minimal effort.
The tool is under continuous development and will be further improved and extended over time.

\begin{table}[!t]
\setlength{\tabcolsep}{3.4pt}
\small
\centering
\caption{Evaluated models with their name, type, parameter size and exact tag.}
\label{tab:models_used}
\begin{tabularx}{\linewidth}{@{\hspace{0pt}}l@{\hspace{5pt}}l@{\hspace{-1pt}}r@{\hspace{3pt}}r@{\hspace{3pt}}X}
\toprule
& \textbf{Name} & \textbf{T\textsuperscript{1}} & \textbf{Size} & \textbf{Tag} \\
\midrule

\multirow{5}{*}{\rotatebox{90}{mid-2024}} 
& Llama-3 & O & 8B & \texttt{\scriptsize meta-llama-3-8b-instruct.Q8\_0} \\
& Llama-3 & O & 70B & \texttt{\scriptsize meta-llama-3-70b-instruct.Q4\_K\_M} \\
& Gemini 1.5 Flash & C & n/a & \texttt{\scriptsize google-gemini-1.5-flash-001} \\
& Claude 3.5 Sonnet & C & n/a & \texttt{\scriptsize anthropic-claude-3-5-sonnet@20240620} \\
& GPT-4o & C & 8B & \texttt{\scriptsize openai-gpt-4o-2024-05-13} \\

\midrule

\multirow{13}{*}{\rotatebox{90}{fall 2025}} 
& Llama 3.1 & O & 8B & \texttt{\scriptsize meta-llama-3.1-8B-instruct-gguf.Q8.0o} \\
& DeepSeek R1 0528 & O & 8B & \texttt{\scriptsize deepseek-r1-0528-qwen3-8b.Q4\_K\_M} \\
& Qwen3 & O & 8B & \texttt{\scriptsize qwen-qwen3-8b} \\
& DeepSeek R1 0528 & O & 685B & \texttt{\scriptsize deepseek-r1-0528} \\
& GPT OSS 20B & O & 20B & \texttt{\scriptsize openai-gpt-oss-20b} \\
& Llama 4 Scout & O & 17B & \texttt{\scriptsize meta-llama-4-scout-17b-16e-instruct} \\
& Llama 3.3 & O & 70B & \texttt{\scriptsize meta-llama-3.3-70b-instruct} \\
& Qwen3 & O & 235B & \texttt{\scriptsize qwen-qwen3-235b-a22b-instruct-2507} \\
& Claude Sonnet 4.5 & C & n/a & \texttt{\scriptsize anthropic-claude-sonnet-4-5-20250929} \\
& Gemini 2.5 Flash & C & n/a & \texttt{\scriptsize google-gemini-2.5-flash} \\
& GPT 5 & C & n/a & \texttt{\scriptsize openai-gpt-5-2025-08-07} \\
& GPT 5 Mini & C & n/a & \texttt{\scriptsize openai-gpt-5-mini-2025-08-07} \\
& GPT 5 Nano & C & n/a & \texttt{\scriptsize openai-gpt-5-nano-2025-08-07} \\

\bottomrule
\end{tabularx}
\scriptsize
\raggedright
\vspace*{0.25em} 

$^1$~Type: open (weights) / commercial
\end{table}

\section{Evaluation}
\label{sec:evaluation}

To assess the applicability of our pipeline for filtering the paper corpus of an SLR using LLM agents under human supervision, we employed data collected for a preliminary version of our recent SLR, 
``Visual Network Analysis in Immersive Environments: A Survey''~\cite{joos2025visualnetworkanalysisimmersive}.
This topic provides a sufficiently large set of potentially relevant papers, as it covers research involving immersive technologies such as virtual reality and augmented reality, with a focus on graph-based data.
Since papers related to this topic often address diverse domains, they are distributed across many journals and conference proceedings.
Consequently, multiple online libraries must be queried to gather the relevant works, which results in a large number of retrieved papers, many of which are not relevant.

\begin{table}[!t]
\setlength{\tabcolsep}{3.4pt}
\small
\centering
\caption{Evaluation results for the \textbf{Mid-2024} LLM agents and the two consensus schemes (all models and top-3 models only) based on our reference survey, using the validated human classification as the ground truth.}
\label{tab:OverallResults}
\begin{tabularx}{\linewidth}{@{\hspace{4pt}}X@{\hspace{0pt}}r|rrrrr|rr}
\toprule
& \rotatebox{0}{Metric}
& \rotatebox{90}{\parbox{1.5cm}{Llama-3\\(8B)}} %
& \rotatebox{90}{\parbox{1.5cm}{Llama-3\\(70B)}}
& \rotatebox{90}{\parbox{1.5cm}{Gemini 1.5\\Flash}}
& \rotatebox{90}{\parbox{1.5cm}{Claude 3.5\\Sonnet}}
& \rotatebox{90}{GPT-4o} 
& \rotatebox{90}{\parbox{1.5cm}{Consensus\\(All)$\,^1$}}
& \rotatebox{90}{\parbox{1.5cm}{Consensus\\(Best)$\,^2$}} \\
\midrule
\multirow{4}{*}{\rotatebox{90}{Counts}} 
& \textbf{TP}$\;(\uparrow)$ & 86 & 85 & 67 & 76 & 80 & \textbf{87} & \textbf{87}\\
& \textbf{FP}$\;(\downarrow)$ & 774 & 194 & 91 & 95 & \textbf{50} & 862 & 167 \\
& \textbf{TN}$\;(\uparrow)$ & 7461 & 8041 & 8144 & 8140 & \textbf{8185} & 7373 & 8068 \\
& \textbf{FN}$\;(\downarrow)$ & 2 & 3 & 21 & 12 & 8 & \textbf{1} & \textbf{1} \\
\midrule
\multirow{4}{*}{\rotatebox{90}{Evaluation}}
& \textbf{Acc.}$\;(\uparrow)$  & 90.68  & 97.63 & 98.65  & 98.71  & \textbf{99.30}  & 89.63 & 97.98 \\
& \textbf{Prec.}$\;(\uparrow)$  & 10.00  & 30.47 & 42.41  & 44.44  & \textbf{61.54}  & 9.17 & 34.25 \\
& \textbf{Rec.}$\;(\uparrow)$  & 97.73  & 96.59 & 76.14  & 86.36  & 90.91  & \textbf{98.86} & \textbf{98.86} \\
& \textbf{F\textsubscript{1}}$\;(\uparrow)$ & 18.14  & 46.32 & 54.47  & 58.69  & \textbf{73.39}  & 16.78 & 50.88 \\
\bottomrule
\end{tabularx}
\scriptsize
\raggedright
\vspace*{0.25em} 

$^1$~Consensus between all (five) models. $\quad$
$^2$~Consensus between the three best-performing models (i.e., without the  Llama3 variants).%
\end{table}

For the initial paper selection, we followed the \textit{PRISMA} methodology~\cite{prisma2009}, beginning with a structured keyword search in paper titles and abstracts within major computer science repositories.
In our case, we included papers from the \textit{ACM Digital Library}, \textit{IEEE Xplore}, and \textit{Eurographics}.
After unifying metadata formats, removing duplicates, and excluding non-paper entries, we obtained an initial corpus of \textbf{8,323} papers in the preliminary version (the published version used a later iteration with additional results).
The papers were then manually screened through several iterations.
This process was highly time-consuming and required multiple researchers over several weeks, but it produced the ground-truth classification: among the initial corpus, \textbf{88} papers were identified as relevant and \textbf{8,235} as irrelevant.
Using this ground-truth dataset, we examined the potential of LLMs to assist in the labor-intensive task of corpus filtering.

We began by testing the LLMs with various prompt formulations, instructing them to classify each paper individually, and quickly identified a prompt structure that performed reliably.
In this structure, the LLM is first informed about its role and task, followed by the required output format and the paper's title and abstract.
For the final version of the prompt (see \autoref{fig:prompt_format} and \ref{sec:appendix}), we introduced additional inclusion and exclusion criteria.

In the following, we present the results we obtained through three different evaluations.
For the first one, we used models that represented the state of the art in mid-2024, including two open and three commercial LLMs.
At that time, the number of openly available models was very limited.
Since then, the situation has changed considerably, as open models from Meta, Deepseek, Alibaba, OpenAI, and others have become available, rivaling commercial systems in performance.
This development offers substantial potential for research applications, particularly where funding constraints and data protection requirements are critical.
Given these rapid advancements in model quality and accessibility, we repeated the evaluation with state-of-the-art models available in fall 2025, consisting of eight open and five commercial LLMs.
Lastly, for one model (\texttt{Llama 3.1 8B}), we investigated the influence of prompt design to explore how its relatively high \textit{false-negative} rate could be reduced.
The results of these three evaluation phases are presented in the following sections.

\subsection{Mid-2024 LLMs}

\begin{figure}[!th]
    \centering
    \includegraphics[width=1\linewidth]{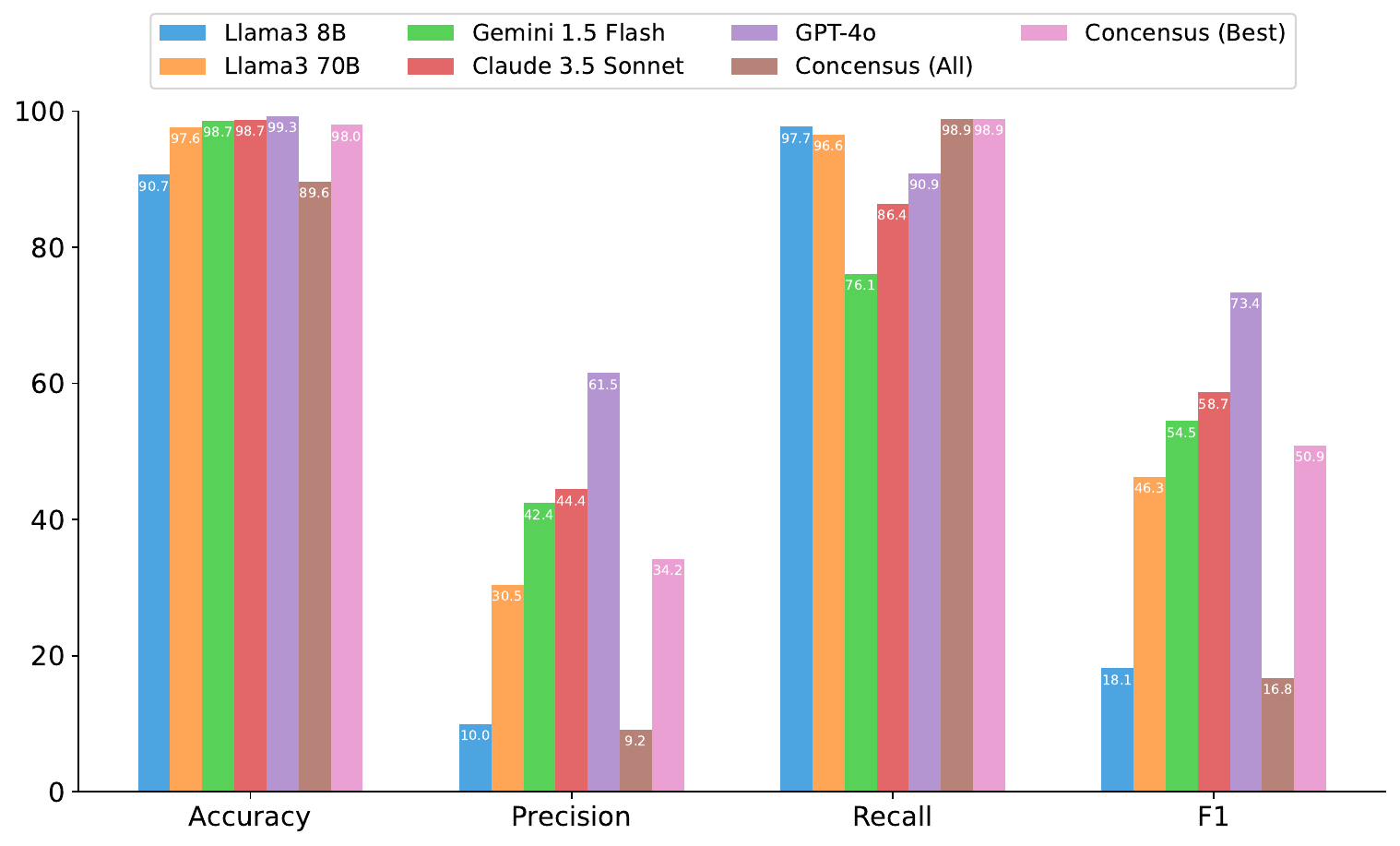}
    \caption{Performance of the \textbf{Mid-2024} LLMs in terms of \textit{accuracy}, \textit{precision}, \textit{recall}, and \textit{F\textsubscript{1} Score}. The open models achieve higher \textit{recall}, while the commercial models perform better in \textit{precision} and \textit{F\textsubscript{1} Score}. Both consensus schemes show very high \textit{recall}, yet the approach using all LLMs produces more \textit{false positives}, which reduces \textit{precision}.}
    \label{fig:llm-metrics}
    \vspace*{-2.3pt}
\end{figure}

Our initial evaluation involved five models that represented the state of the art among LLMs in mid-2024.
These included two open Llama models from Meta, one smaller and one larger, as well as the leading commercial models developed by Google, Anthropic, and OpenAI.
The specific models are identified in detail in the top part of \autoref{tab:models_used}.

The results of the individual LLM classifications are summarized in \autoref{tab:OverallResults}.
Overall, the models performed well, achieving accuracies above 90~\% across all cases (see \autoref{fig:llm-metrics}).
Nevertheless, some clear differences can be observed. The open models, particularly \texttt{Llama3 8B}, adopted a more conservative strategy, tending to include more papers overall. This led to a higher number of false positives (FP) but fewer false negatives (FN), as these models aimed to avoid excluding relevant papers.
In contrast, the commercial models were stricter, discarding a larger share of papers, which increased their true negative (TN) rates but also resulted in more false exclusions (higher FN rates).

Interestingly, the incorrectly classified papers varied considerably between models.
For erroneous \textit{inclusions} (FP), in most cases only one model, which was most often \texttt{Llama3 8B}, was responsible for the misclassification (see \autoref{fig:llm-inclusion-exclusion}, left).
The number of papers falsely included by multiple models was significantly lower.
A similar pattern emerged for relevant papers that were incorrectly excluded (FN), where individual models (most notably \texttt{Gemini 1.5 Flash}) were usually responsible for the mistakes, while overlaps among multiple models were rare (see \autoref{fig:llm-inclusion-exclusion}, right).

\begin{table*}[!th]
\setlength{\tabcolsep}{2.5pt}
\small
\centering
\caption{Evaluation results comparing the classifications produced by the \textbf{Fall 2025} LLM agents (13 in total) and the five consensus schemes with the human-generated ground truth. The best performance is achieved by the consensus combining GPT-5, Claude Sonnet 4.5, and Llama 3.3 (70B) models (\textit{Consensus Best}).}
\label{tab:new-OverallResults}

\begin{tabularx}{\linewidth}{
  l|
  *{13}{>{\raggedleft\arraybackslash}X}|
  *{5}{>{\raggedleft\arraybackslash}X}
}
\toprule
Metric & \rotatebox{90}{Llama 3.1 8B}
& \rotatebox{90}{\parbox{2cm}{DeepSeek R1\\0528 8B}}
& \rotatebox{90}{Qwen3 8B}
& \rotatebox{90}{\parbox{2cm}{DeepSeek R1\\ 0528}}
& \rotatebox{90}{GPT OSS 20B}
& \rotatebox{90}{Llama 4 Scout}
& \rotatebox{90}{Llama 3.3 70B}
& \rotatebox{90}{Qwen3}
& \rotatebox{90}{\parbox{2cm}{Claude\\ Sonnet 4.5}}
& \rotatebox{90}{\parbox{2cm}{Gemini 2.5\\Flash}}
& \rotatebox{90}{GPT 5}
& \rotatebox{90}{GPT 5 Mini}
& \rotatebox{90}{GPT 5 Nano}
& \rotatebox{90}{\parbox{2cm}{Consensus\\(Open 8B)$\,^1$}}
& \rotatebox{90}{\parbox{2cm}{Consensus\\(Open Large)$\,^2$}}
& \rotatebox{90}{\parbox{2cm}{Consensus\\(Commercial)$\,^3$}}
& \rotatebox{90}{\parbox{2cm}{Consensus\\(Best)$\,^4$}}
& \rotatebox{90}{\parbox{2cm}{Consensus\\ (All)$\,^5$}} \\
\midrule
\textbf{TP}$\;(\uparrow)$  & 72 & 80 & 56 & 77 & 77 & \textbf{86} & 85 & 72 & 77 & 79 & 80 & 85 & 84 & 82 & \textbf{88} & \textbf{88} & \textbf{88} & \textbf{88} \\
\textbf{FP}$\;(\downarrow)$ & 98 & 386 & 33 & 68 & 58 & 341 & 140 & \textbf{30} & 39 & 68 & 58 & 134 & 135 & 426 & 382 & 183 & \textbf{166} & 653 \\
\textbf{TN}$\;(\uparrow)$  & 8137 & 7849 & 8202 & 8167 & 8177 & 7894 & 8095 & \textbf{8205} & 8196 & 8167 & 8177 & 8101 & 8100 & 7809 & 7853 & 8052 & \textbf{8069} & 7582 \\
\textbf{FN}$\;(\downarrow)$ & 16 & 8 & 32 & 11 & 11 & \textbf{2} & 3 & 16 & 11 & 9 & 8 & 3 & 4 & 6 & \textbf{0} & \textbf{0} & \textbf{0} & \textbf{0} \\
\midrule
\textbf{Acc.}$\;(\uparrow)$  & 98.63 & 95.27 & 99.22 & 99.05 & 99.17 & 95.88 & 98.28 & \textbf{99.45} & 99.40 & 99.07 & 99.21 & 98.35 & 98.33 & 94.81 & 95.41 & 97.80 & \textbf{98.01} & 92.15 \\
\textbf{Prec.}$\;(\uparrow)$ & 42.35 & 17.17 & 62.92 & 53.10 & 57.04 & 20.14 & 37.78 & \textbf{70.59} & 66.38 & 53.74 & 57.97 & 38.81 & 38.36 & 16.14 & 18.72 & 32.47 & \textbf{34.65} & 11.88 \\
\textbf{Rec.}$\;(\uparrow)$  & 81.82 & 90.91 & 63.64 & 87.50 & 87.50 & \textbf{97.73} & 96.59 & 81.82 & 87.50 & 89.77 & 90.91 & 96.59 & 95.45 & 93.18 & \textbf{100.0} & \textbf{100.0} & \textbf{100.0} & \textbf{100.0} \\
\textbf{F\textsubscript{1}}$\;(\uparrow)$ & 55.81 & 28.88 & 63.28 & 66.09 & 69.06 & 33.40 & 54.31 & \textbf{75.79} & 75.49 & 67.23 & 70.80 & 55.37 & 54.72 & 27.52 & 31.54 & 49.03 & \textbf{51.46} & 21.23 \\
\bottomrule
\end{tabularx}
\scriptsize
\raggedright
\vspace*{0.25em} 

$^1$~Consensus between all small (8B) open models (3): \texttt{Llama 3.1 8B, DeepSeek R1
0528 8B, Qwen3 8B} $\quad$
$^2$~Consensus between all large ($>$8B) open models ($N=5$): \texttt{DeepSeek R1 0528, GPT OSS 20B, Llama 4 Scout, Llama 3.3 70B, Qwen3} $\quad$
$^3$~Consensus between all commercial models ($N=5$): \texttt{Claude Sonnet 4.5, Gemini 2.5 Flash, GPT 5 (Full/Mini/Nano)} $\quad$
$^4$~Consensus between best combination of models ($N=3$): \texttt{GPT 5, Claude Sonnet 4.5, Llama 3.3 70B}. $\quad$
$^5$~Consensus between all models ($N=13$). $\quad$
\end{table*}

To further explore this, we analyzed the performance of consensus-based voting strategies: one combining all LLMs (\textit{Consensus (All)}) and another including only the three best-performing models (\textit{Consensus (Best)}), namely \texttt{Gemini 1.5 Flash}, \texttt{Claude 3.5 Sonnet}, and \texttt{GPT-4o}.
In these consensus approaches, a paper is discarded only if all participating LLMs agree to exclude it, and it is included if at least one model suggests inclusion.
The outcomes of both consensus methods (see \autoref{tab:OverallResults}, right columns) are highly promising, showing excellent true positive (TP) and false negative (FN) rates.
Only one relevant paper would have been excluded, and a manual review revealed that this case was ambiguous even for the human evaluators.
While both consensus approaches achieved identical TP and FN rates, which are the most important for our use case, the \textit{Consensus (Best)} method showed a notably lower false positive rate (see \autoref{fig:matrix}).
This reduced the number of papers requiring manual filtering by 695 and required only three instead of five models, offering considerable improvements in both efficiency and cost.

In summary, the mid-2024 models demonstrate strong performance, achieving accuracies above 90~\% in paper classification.
However, the commercial models still exhibit relatively high false negative rates.
The results suggest that more complex models tend to discard a larger number of papers, which reduces the false positive rate but increases the false negative rate.
Consequently, these models benefit particularly from the consensus approach.

\subsection{Fall 2025 LLMs}

We repeated the original study more than one year later, incorporating state-of-the-art models available in fall 2025.
With the continuous growth of open LLMs, we included eight such models in this evaluation. 
Alongside three different \texttt{Llama} variants developed by Meta, the set comprised two DeepSeek models (\texttt{R1 0528}), two \texttt{Qwen3} models from Alibaba, and the open variant provided by OpenAI (\texttt{GPT OSS 20B}).
Among these, three are relatively small models with eight billion parameters, selected because they can operate on basic hardware setups, enhancing accessibility and practical use for researchers.

In addition to the open models, we evaluated five commercial LLMs.
As in the previous study, these included the latest releases from Anthropic (\texttt{Claude Sonnet 4.5}), Google (\texttt{Gemini 2.5 Flash}), and OpenAI (\texttt{GPT 5}).
We also incorporated two smaller OpenAI variants, \texttt{GPT 5 Mini} and \texttt{GPT 5 Nano}, which offer faster performance and reduced computational cost.
The specific models are summarized in the bottom part of \autoref{tab:models_used}.

\begin{figure*}
    \centering
    \begin{subfigure}{\textwidth}
        \includegraphics[width=\textwidth]{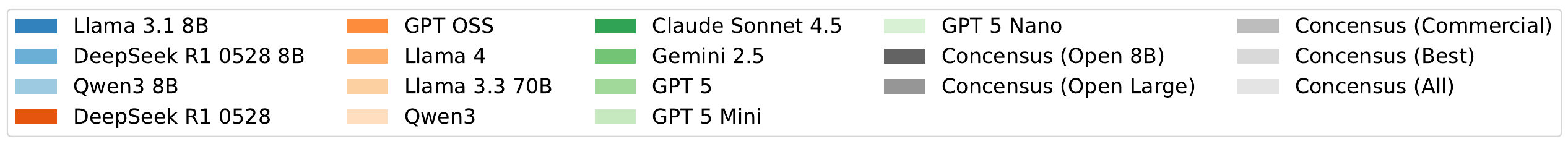}
    \end{subfigure}
    
    \begin{subfigure}{0.49\textwidth}
        \includegraphics[width=\textwidth]{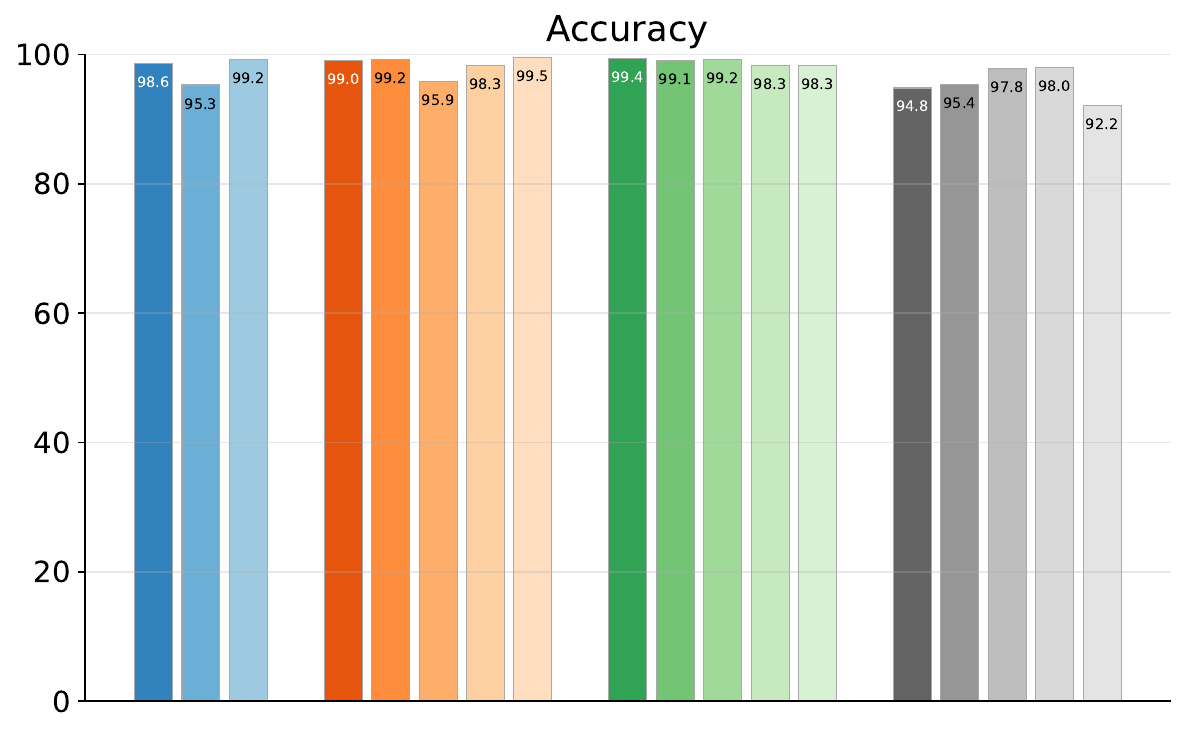}
    \end{subfigure}
    \hfill
    \begin{subfigure}{0.49\textwidth}
        \includegraphics[width=\textwidth]{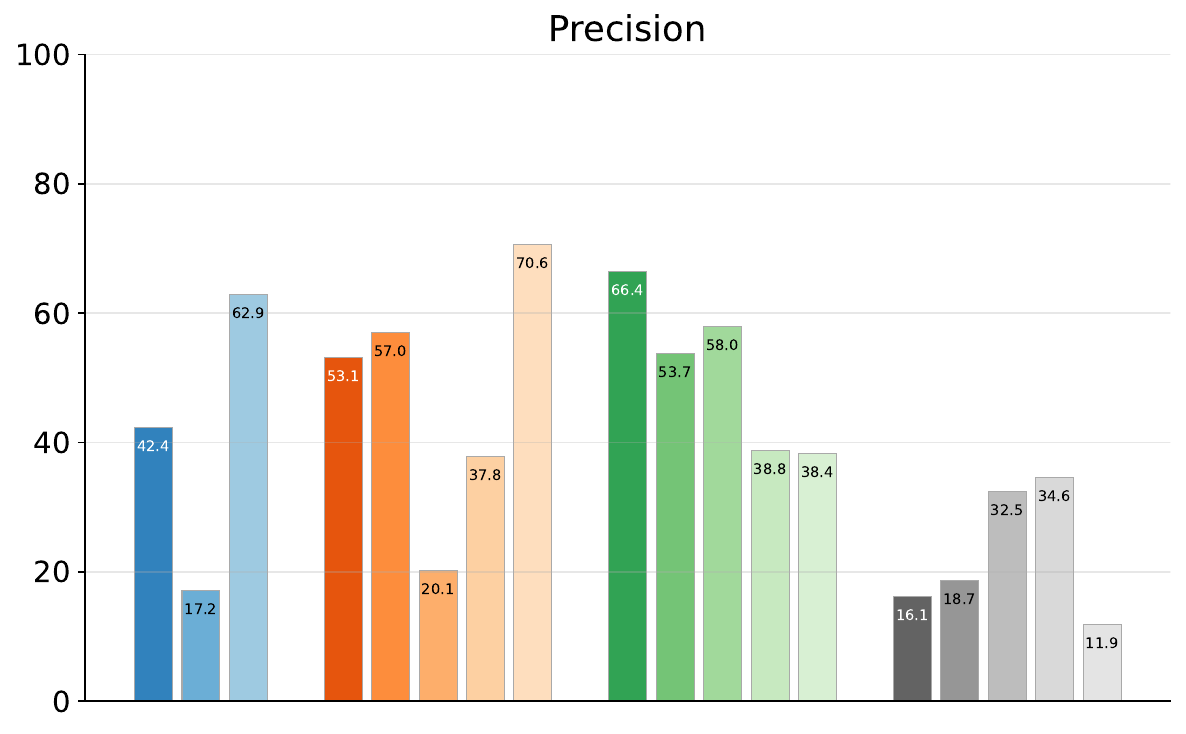}
    \end{subfigure}
    
    \begin{subfigure}{0.49\textwidth}
        \includegraphics[width=\textwidth]{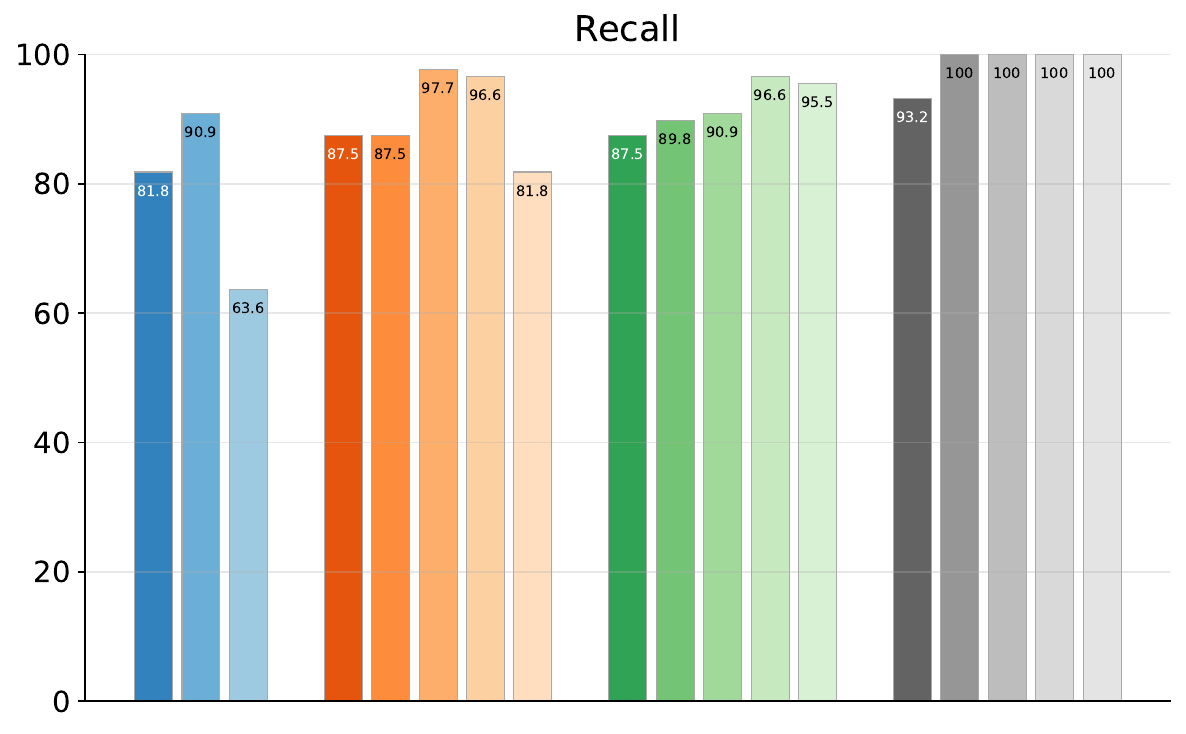}
    \end{subfigure}
    \hfill
    \begin{subfigure}{0.49\textwidth}
        \includegraphics[width=\textwidth]{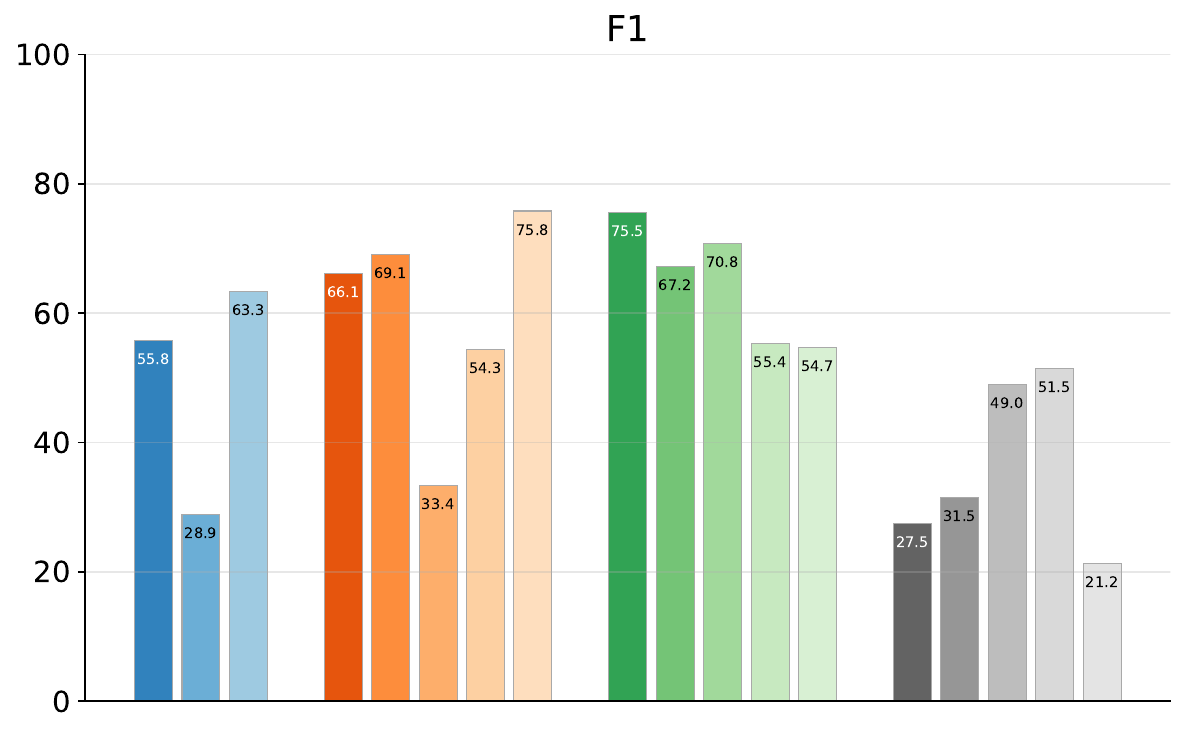}
    \end{subfigure}
    
    \caption{Performance of the \textbf{Fall 2025} LLMs in regarding \textit{accuracy}, \textit{precision}, \textit{recall}, and \textit{F\textsubscript{1} Score}. All models achieve high \textit{accuracy}, while the commercial and large open models particularly excel in \textit{recall}. The small and large Qwen3 models exhibit noticeable weaknesses in this metric. All consensus schemes, except for \textit{Open 8B}, successfully include all relevant papers.}
    \label{fig:new-llm-metrics}
\end{figure*}

The results of the individual LLM classifications are summarized in \autoref{tab:new-OverallResults}.
All models followed the prompt and provided clear decisions for each paper, except for \texttt{DeepSeek R1 0528 8B}, which caused issues in 34 cases by returning incomplete answers or irrelevant information.
In these instances, the papers were included, as the primary objective is to avoid losing relevant items.
The progress and refinement of LLMs over the past year are clearly reflected in the outcomes.
All models achieved higher accuracies, ranging from approximately 95~\% to 99~\% (see \autoref{fig:new-llm-metrics}).
Similarly, both precision and F\textsubscript{1} scores improved considerably, especially for the 8B variant of the \texttt{Llama} model.
While the mid-2024 version (\texttt{Llama 3 8B}) incorrectly included 774 papers, the newer fall 2025 model (\texttt{Llama 3.1 8B}) reduced this number to only 98 false positives.
This trend is consistent across most models, which generally became more selective and produced fewer false positives.

\begin{figure*}[!th]
    \centering
    \includegraphics[width=1\linewidth]{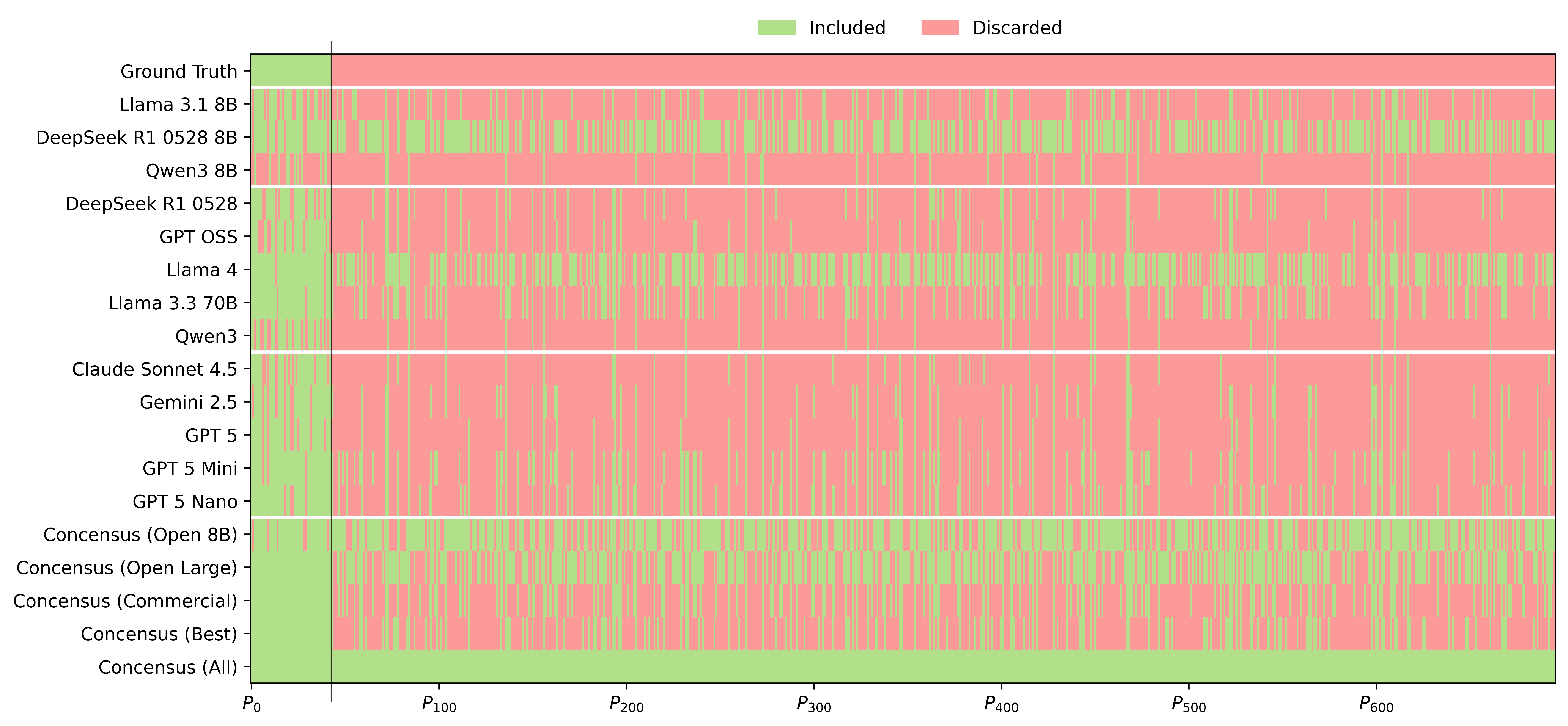}
    \caption{Overview of pairwise comparisons illustrating \emph{incorrect} decisions made by the \textbf{Fall 2025} models. The ground-truth classification of each paper is displayed in the top row (left side: included, right side: discarded). The subsequent rows present the results for the small open models (8B), larger open models, commercial models, and five distinct consensus schemes. Incorrect exclusions (FN) are shown as \textcolor{colorDiscarded}{red discarded} lines on the left, while incorrect inclusions (FP) are shown as \textcolor{colorIncluded}{green included} lines on the larger right side. For most consensus schemes, except the one involving only small open models, no relevant papers are lost.}
    \label{fig:new-matrix}
\end{figure*}

\begin{figure*}
    \centering
    \begin{subfigure}[b]{0.49\textwidth}
        \centering
        \includegraphics[width=\textwidth]{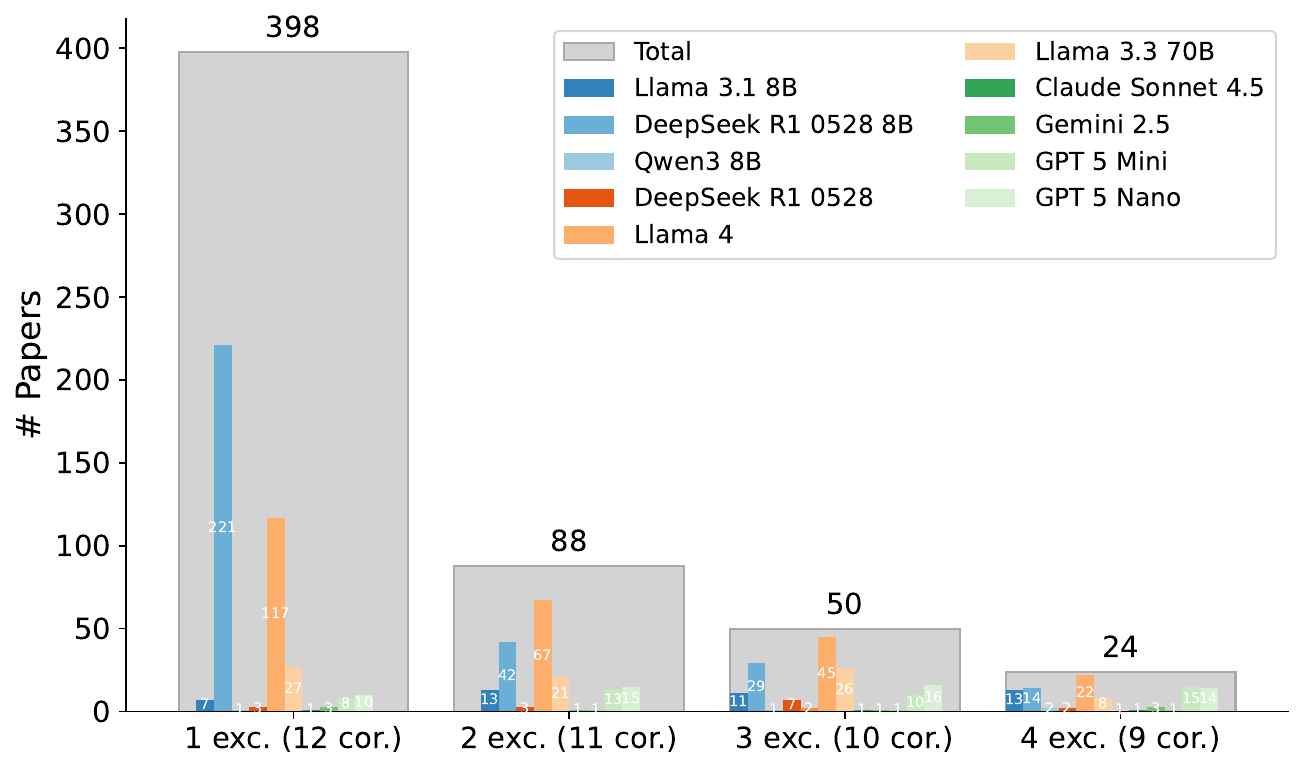}
    \end{subfigure}
    \hfill
    \begin{subfigure}[b]{0.49\textwidth}
        \centering
        \includegraphics[width=\textwidth]{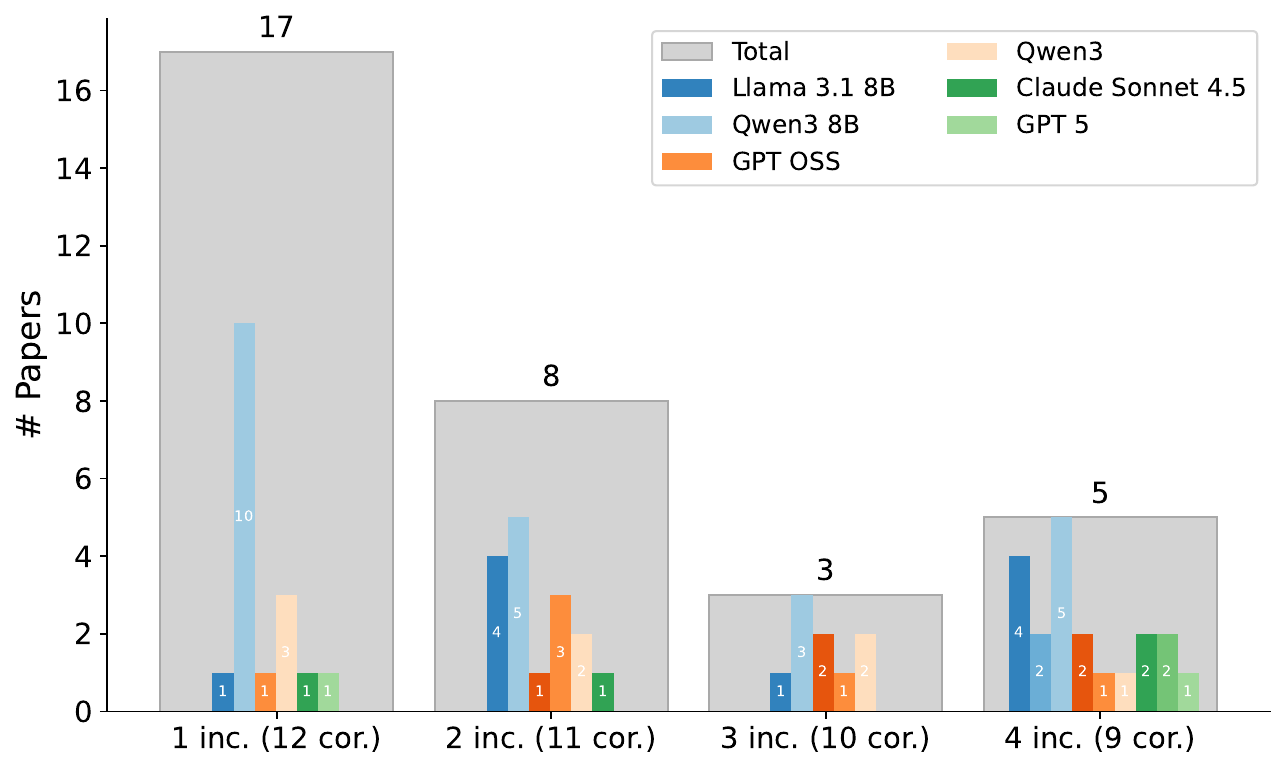}
    \end{subfigure}
    \caption{Number of papers (gray background) that were \emph{incorrectly (inc.)} classified as \textbf{included} (left) or \textbf{excluded} (right) by the \textbf{Fall 2025} LLM agents, grouped by the number of agents involved in each incorrect decision (up to four). The individual bars show how often a specific agent contributed to an incorrect classification. It can be observed that \texttt{Llama 3.1 8B} and \texttt{Llama 4 Scout} frequently include papers falsely that other LLMs do not, while \texttt{Qwen3 8B} is often responsible (along with a few other models) for incorrect exclusions.}
    \label{fig:new-llm-inclusion-exlusion}
\end{figure*}

Regarding false inclusions of papers, especially \texttt{DeepSeek R1 0528 8B} and \texttt{Llama 4 Scout} tended to classify irrelevant papers as relevant against the vote of the remaining models (see \autoref{fig:new-llm-inclusion-exlusion} left).
The fall 2025 models exhibited slightly higher false negative rates compared with the earlier models from mid-2024.
In particular, the \texttt{Qwen3} models tended to exclude relevant papers that most other LLMs included (see \autoref{fig:new-llm-inclusion-exlusion}, right).
Nonetheless, the larger \texttt{Qwen3} variant achieved the highest overall accuracy, precision, and F\textsubscript{1} score, though this came at the cost of 16 falsely discarded papers.

Given the stricter decision patterns of the fall 2025 models, which result in reduced recall, the consensus approach becomes especially valuable for the final classification (see \autoref{fig:new-matrix}).
Unlike in the previous evaluation, a consensus that includes all models (\textit{Consensus All}) successfully retained all relevant papers.
However, this approach also led to over 600 false positives, increasing the manual review workload.
Smaller subsets of models achieved better balance.
For example, combining only the larger open models (\textit{Consensus Open Large}) resulted in 382 false positives, while using only commercial models (\textit{Consensus Commercial}) reduced the number to 183.
The best overall result was obtained by combining three models, namely \texttt{GPT-5}, \texttt{Claude Sonnet 4.5}, and \texttt{Llama 3.3 (70B)}, in the \textit{Consensus Best} scheme, which produced only 166 false positives and no false negatives.
The consensus composed solely of small open models was the only one that excluded six relevant papers, possibly due to their limited reasoning capabilities or the need for more refined prompt designs.
This motivated further exploration of prompt optimization for smaller open models, as discussed in \autoref{sec:prompt-opt}.

In summary, the results demonstrate clear progress in LLM performance over the past year, with significantly improved accuracy and precision.
Open models, in particular, now show strong performance and accessibility, making them highly suitable for research contexts.
A combination of open models correctly classified all relevant papers while including only slightly more than 300 irrelevant ones, substantially reducing the manual workload.
These findings highlight the robustness of the consensus approach, where individual model errors are often compensated by others, resulting in more stable and reliable outcomes.

\subsection{Prompt Optimization}
\label{sec:prompt-opt}

The use of small open models is typically more accessible and affordable for researchers than relying on large-scale models that demand substantial GPU resources or commercial solutions, which can be costly and raise concerns about data privacy.
However, our evaluations revealed that small models (i.e., around 8B) still struggle to ensure that no relevant papers are excluded, even when combining three of them in a consensus scheme.
To address this limitation, we investigated whether refining the prompt could help mitigate the issue.
We selected the \texttt{Llama 3.1 8B} model for this analysis, as it produced more balanced results than the other two open 8B models.
In total, this model misclassified 16 papers that should have been included as irrelevant.

To identify the cause of these misclassifications, we examined the model's reasoning, which was requested as part of the original prompt.
For one paper, the model produced the following response:
\begin{quote}
\textit{The paper focuses on ``graph visualization'' using alternative display environments like spatially-aware mobile devices and a large display wall, but it lacks explicit mention of immersive environments like virtual reality, augmented reality, or extended reality. While it discusses interaction modalities beyond traditional desktop interfaces, the context is more about non-immersive multi-display settings rather than immersive environments}
\end{quote}

For another relevant paper, the model reasoned: 
\begin{quote}
\textit{This paper deals with genomic data and uses virtual reality, but it does not mention network visualization as its primary focus. The abstract suggests that a node-link diagram is used for visualizing gene expression networks, however, it's more focused on the application of VR in genomics than on ``visual network analysis''}
\end{quote}

In both examples (and in many similar cases), the model correctly identified that networks were visualized in immersive contexts as defined in the prompt.
Nevertheless, it considered the applications too specific or the focus not sufficiently centered on network analysis, leading to incorrect exclusions.
Since the model extracted the relevant information correctly but applied overly strict inclusion rules, we designed seven alternative prompts to relax these constraints while avoiding a substantial increase in false positives.
The general concepts behind these prompts are summarized below, while the full prompt texts and modifications are provided in \ref{sec:appendix}.

\paragraph{$P_{orig}$}  
In the original prompt, we defined the model's role and task, described the topic with examples, and provided explicit inclusion and exclusion criteria along with a structured answer format.
The goal of this initial version was to minimize false positives, instructing the model to discard papers whenever it was uncertain about their relevance.

\paragraph{$P_1$}  
In the first modification, we reversed the uncertainty rule, instructing the model to include papers when unsure about their relevance.
This adjustment aimed to reduce the false negatives.

\paragraph{$P_2$}  
The second prompt built upon $P_1$ and added context about the purpose of the task.
We explicitly told the model that it served as a first filtering step and that false negatives were more problematic than false positives, lowering the threshold for inclusion by clarifying the actual research goal.

\paragraph{$P_3$}  
In the third version, we removed the explicit inclusion and exclusion criteria and replaced them with a general instruction to discard a paper only if the model was certain that it was irrelevant, and to include it otherwise.
This aimed to give the model more flexibility and encourage broader reasoning about potential relevance.

\paragraph{$P_4$}  
The fourth prompt combined the instructions from $P_3$ with the task explanation introduced in $P_2$, evaluating how these two strategies together would influence the model's decisions.

\paragraph{$P_5$}  
The fifth prompt extended $P_4$ by clarifying that the goal was not to include all papers indiscriminately but to maintain a balance between reducing false negatives and limiting the manual filtering workload.
This modification encouraged the model to remain cautious without becoming overly restrictive.

\paragraph{$P_6$}  
For the sixth prompt, we asked \texttt{GPT 5} to revise our original prompt to make it more effective for similar LLMs.
The resulting version retained much of the original structure but added more detailed explanations and three explicit examples (two papers to be discarded and one to be included).
It also specified that in cases of uncertainty, a paper should be discarded.

\paragraph{$P_7$}  
The final prompt used the same structure as $P_6$ but modified the ambiguity rule, instructing the model to include papers when uncertain rather than discard them.\\

\begin{table}
\setlength{\tabcolsep}{2.5pt}
\small
\centering
\caption{Results for \texttt{Llama 3.1 8B} using the original prompt $P_{orig}$ and seven alternatives ($P_1$ to $P_7$, see \ref{sec:appendix}). While $P_{orig}$ performs best in \textit{accuracy}, \textit{precision}, and \textit{F\textsubscript{1} Score}, the \textit{recall} is substantially higher for $P_1$ to $P_7$, though at the cost of more \textit{false positives}.}
\label{tab:new-prompt-results}
\begin{tabularx}{\linewidth}{Xr|rrrrrrrr}
\toprule
& \rotatebox{0}{Metric}
& \rotatebox{0}{$P_{orig}$}
& \rotatebox{0}{$P_1$}
& \rotatebox{0}{$P_2$}
& \rotatebox{0}{$P_3$}
& \rotatebox{0}{$P_4$}
& \rotatebox{0}{$P_5$}
& \rotatebox{0}{$P_6$}
& \rotatebox{0}{$P_7$} \\
\midrule
\multirow{4}{*}{\rotatebox{90}{Counts}} 
& \textbf{TP}$\;(\uparrow)$ & 72 & 80 & 82 & 87 & 87 & 87 & 81 & \textbf{88} \\
& \textbf{FP}$\;(\downarrow)$ & \textbf{98} & 176 & 215 & 1690 & 1688 & 1789 & 475 & 3956 \\
& \textbf{TN}$\;(\uparrow)$ & \textbf{8137} & 8059 & 8020 & 6545 & 6547 & 6446 & 7760 & 4279 \\
& \textbf{FN}$\;(\downarrow)$ & 16 & 8 & 6 & 1 & 1 & 1 & 7 & \textbf{0} \\
\midrule
\multirow{4}{*}{\rotatebox{90}{Evaluation}}
& \textbf{Acc.}$\;(\uparrow)$  & \textbf{98.63} & 97.79 & 97.34 & 79.68 & 79.71 & 78.49 & 94.21 & 52.47 \\
& \textbf{Prec.}$\;(\uparrow)$  & \textbf{42.35} & 31.25 & 27.61 & 4.90 & 4.90 & 4.64 & 14.57 & 2.18 \\
& \textbf{Rec.}$\;(\uparrow)$  & 81.82 & 90.91 & 93.18 & 98.86 & 98.86 & 98.86 & 92.05 & \textbf{100.0} \\
& \textbf{F\textsubscript{1}}$\;(\uparrow)$ & \textbf{55.81} & 46.51 & 42.60 & 9.33 & 9.34 & 8.86 & 25.16 & 4.26 \\
\bottomrule
\end{tabularx}
\end{table}

The results obtained with \texttt{Llama 3.1 8B} using the original prompt $P_{orig}$ and the alternative prompts $P_1$--$P_7$ are summarized in \autoref{tab:new-prompt-results}.
While there are major differences in the results, all modified versions led to a substantial increase in recall.
A simple adjustment of the rule for uncertain cases ($P_1$) already reduced the number of false negatives by 50~\%, but also doubled the number of false positives.
Adding a contextual explanation of the task ($P_2$) further decreased the false negatives while causing only a slight increase in false positives.
A more pronounced change occurred when the explicit inclusion and exclusion criteria were replaced by a general instruction ($P_3$).
In this case, only one paper was wrongly excluded, but nearly 1,700 papers were falsely included.
The results for prompts $P_4$ and $P_5$, both derived from $P_3$, were similar.
The prompt modified by the \texttt{GPT 5} model ($P_6$) reduced the number of false negatives from 16 to 7, yet increased the number of false positives to 475.
Finally, changing the rule for uncertain cases in $P_7$ resulted in no false negatives anymore, but this came with a steep trade-off: almost 4,000 papers were wrongly classified as relevant.

In summary, our findings show that the prompt strongly affects the classification results and that even minor modifications can have a substantial impact.
This emphasizes the importance of carefully refining prompts and analyzing their outcomes, as supported by our pipeline and visual-interactive application.
While the modified prompts successfully reduced the number of falsely excluded papers, they also caused a considerable increase in false positives.
This outcome may stem from the limited reasoning capabilities of smaller LLMs, but it also highlights the importance of continued research on prompt design and its influence on model behavior.

\section{Discussion}
\label{sec:discussion}

The results presented in this work demonstrate that LLMs can significantly improve the efficiency of the literature filtering process in SLRs while maintaining a high level of reliability.
Through the evaluation of both open and commercial models across different time periods, we observed substantial progress in the capabilities of models and their accessibility, especially regarding open models.
The comparison between the mid-2024 and fall 2025 evaluations highlights a remarkable evolution of model quality.
Within a single year, the accuracy and precision of open models improved to the point where they matched or even surpassed some commercial systems.
This progress demonstrates that the LLM ecosystem is becoming more inclusive, as open-source alternatives now provide competitive results without high computational or financial cost.
However, the smaller open models are still not at the level of larger and commercial models, which can only be partly mitigated by prompt engineering.
The results also reveal that individual models show distinct behavioral patterns.
Some tend to classify conservatively, preferring inclusion to avoid false negatives, while others exhibit stricter exclusion policies that enhance precision but lower recall.
This diversity of behavior strengthens the rationale for using multiple models in a consensus scheme.
By combining outputs from heterogeneous LLMs, individual weaknesses are compensated and the overall result becomes more stable and transparent.
The experiments confirm that such consensus approaches reduce the likelihood of erroneous exclusions, which is of the highest importance in the early stages of an SLR.

Our findings also show that human supervision and model diversity are essential to achieve dependable outcomes, confirming that human-AI collaboration remains the most effective strategy for responsible automation in research workflows~\cite{Fischer.EthicalAwareness.2022}.
The consensus-based classification proved to be a robust approach for balancing recall and precision.
When the goal is to avoid missing relevant papers, it is acceptable to include some irrelevant ones, as subsequent manual review can efficiently remove them.
Our best-performing consensus of three models (fall 2025) reached perfect recall with only a moderate increase in false positives (166).
In contrast, the human error range varies depending on factors such as task difficulty, familiarity, stress, and repetition~\cite{Smith.HumanErrorRates.2011}.
Frameworks like HEART~\cite{humphreys1988human}, TESEO~\cite{bello1980human}, and THERP~\cite{kirwan1988comparative} indicate error rates between 0.5~\% and 9~\%, which are comparable to or higher than the error rates of the LLMs (and their consensus) in our study.

The experiments further indicate that using a large number of models in consensus voting can slightly increase false positives, while smaller subsets of well-chosen models maintain high recall with reduced redundancy.
Therefore, researchers may prefer a selective combination of high-performing models rather than an all-inclusive ensemble.
The design of consensus rules can also adapt dynamically based on the stage of the review process.
For example, early iterations can prioritize recall, whereas later iterations can aim for higher precision after the corpus has been substantially reduced.

The prompt optimization study for a small open model revealed that small modifications in task instructions can lead to large variations in model behavior.
Even minor rule adjustments can significantly shift the trade-off between recall and precision.
This sensitivity illustrates both the strength and the fragility of prompt-based control.
While flexible, it requires careful calibration and contextual understanding by the user.
Automated pipelines cannot fully replace human insight in this process.
Interactive supervision through the \textit{LLMSurver} interface proved highly valuable for this reason.
By allowing users to inspect model reasoning and adjust prompts iteratively, the system enables an informed refinement cycle that stabilizes results over time.
This feedback loop is essential for building trust in semi-automated screening systems and ensures that the final outcomes remain transparent and reproducible.

The broader implication of this work is that LLM-based systems can transform how researchers conduct systematic reviews.
Instead of replacing human expertise, they function as intelligent assistants that accelerate routine tasks and allow researchers to focus on higher-level synthesis and interpretation.
By leveraging open models and transparent visual analytics, the process remains accountable and accessible, supporting both reproducibility and equity in scientific research.
The ability to rapidly filter large corpora without sacrificing reliability opens new possibilities for maintaining up-to-date reviews in fast-moving fields where manual approaches are not feasible.

\subsection{Limitations and Future Directions}
\label{sec:limitations_future_work}

Our study has several limitations that suggest directions for future work.
First, the evaluation focused on a single topic and dataset within computer science, and its generalization to other research areas remains to be confirmed. Different fields may employ more ambiguous terminology or less structured abstracts, which could pose challenges for current LLM reasoning capabilities.
We did not use LLMs for identifying candidate papers at the initial stage to avoid the risk of erroneous references~\cite{hadi2023survey}. Exploring this application could nevertheless provide valuable insights. Moreover, since our approach relies on abstract-level information, it does not include full-text context, which can affect inclusion decisions. Future versions of the pipeline could incorporate full-text or citation data to mitigate this limitation.
Our evaluation also indicates that the models themselves evolve continuously, making reproducibility across updates uncertain. While the consensus approach helps reduce false exclusions, it cannot remove them entirely, so manual verification remains an essential safeguard. When additional mechanisms such as \textit{snowballing}~\cite{wohlin2014guidelines} are used, a small number of false exclusions may be acceptable given the substantial efficiency gains.
Finally, we did not conduct a formal user study of \textit{LLMSurver}. Such an evaluation could provide valuable insights into usability and guide further development.

Several promising directions arise from this work. Integrating access to online libraries for automatic corpus retrieval could accelerate the process. Adaptive consensus methods that adjust to model confidence rather than relying on fixed voting thresholds may improve precision. Expanding \textit{LLMSurver} into a collaborative platform where multiple reviewers can jointly inspect, annotate, and discuss results would enhance transparency and support reproducibility. Beyond corpus filtering, similar pipelines could be adapted for related academic tasks, including content screening and snowballing.

\section{Conclusion}
\label{sec:conclusion}

This work presents a semi-automated pipeline that leverages large language models to accelerate and enhance literature filtering for systematic literature reviews.
By combining multiple LLMs in a consensus scheme and integrating human supervision through our open-source tool \href{https://github.com/dbvis-ukon/LLMSurver}{{\color{linkColor}LLMSurver}}, researchers can efficiently reduce large corpora while maintaining high recall and transparency.
Our evaluation using more than 8,000 papers shows that modern fall 2025 models, both open and commercial, achieve performance comparable to or exceeding human screening, reducing weeks of manual effort to minutes.
The rapid evolution of open models over the past year highlights how accessible, cost-effective, and privacy-preserving open AI tools can now match proprietary systems in quality for this task.
Furthermore, our results underscore the critical role of the human supervising the results and modifying the process, as small modifications, for instance, in the prompt, can strongly affect the outcome.
Overall, this study demonstrates that LLM-assisted and consensus-based workflows controlled through human-AI collaboration can streamline and facilitate academic work.

\section*{Declaration of competing interest}
The authors declare that they have no known competing financial interests or personal relationships that could have appeared to
influence the work reported in this paper.

\section*{Declaration of generative AI and AI-assisted technologies in the manuscript preparation process}
During the preparation of this work the authors used GPT 5 (OpenAI) in order to improve the language and grammar of the text originally created by the authors. After using this tool/service, the authors reviewed and edited the content as needed and take full responsibility for the content of the published article.
Other uses of generative AI for research purposes (such as the case study involving multiple LLM agents) are detailed in the manuscript.

\appendix
\section{LLM Prompts}
\label{sec:appendix}

\subsection*{\textbf{Original Prompt} ($P_{orig}$)}

You are a professor in computer science conducting a literature review.
Please decide and classify if the following paper belongs to a specific research direction or not.
For this you are provided with the title and the abstract, which should give you sufficient information for a informed and accurate decision.
\\\\
The research direction is the topic of ``visual network analysis in immersive environments''.
\\\\
Therefore include papers that deal with BOTH network AND immersive aspects.
\\\\
Example of network aspects are:

\begin{itemize}[leftmargin=1.2em]
\setlength\itemsep{-0.2em}
    \item network data
    \item graph data
    \item node-link diagrams
\end{itemize}
Examples of immersive environments are:

\begin{itemize}[leftmargin=1.2em]
\setlength\itemsep{-0.2em}
\item virtual reality (VR)
\item augmented reality (AR)
\item mixed reality (MR)
\item extended reality (XR),
\item wall-sized displays, or CAVE setups
\end{itemize}
You MUST discard papers papers that discuss only networks but not immersive environments or only immersive environments.\\
You MUST discard papers about unrelated networks like 5G, mobile phones, computer networks per se, etc.\\
You MUST discard papers that mention networks but do not actually visualize networks/graphs/node-link data (for example they use a high speed network, network connections, etc.).\\
You MUST be thorough and DISCARD papers that do not explicity mention network visualization. If it remains vague or unclear if network data is actually visualized DISCARD the paper.\\
You MUST include papers describing e.g. a study, where a node-link diagram is viewed in virtual reality.
\\\\
Below is the title and abstract. You must only answer with INCLUDE or DISCARD and a 2 sentence reason of why.
\\ \ 
\noindent\rule{\linewidth}{0.5pt} 

\subsection*{\textbf{Prompt 1} ($P_1$), Changes to $P_{orig}$}

{
\color{gray}
\noindent
\verb|[...]|\\
You MUST be thorough and DISCARD papers that do not explicity mention network visualization. If it remains vague or unclear if network data is actually visualized} \highlight{INCLUDE} {\color{gray}the paper.
\\
\verb|[...]|
}
\\ \ 
\noindent\rule{\linewidth}{0.5pt} 

\subsection*{\textbf{Prompt 2} ($P_2$), Changes to $P_{orig}$}

{
\color{gray}
\noindent
\verb|[...]|\\
You MUST be thorough and DISCARD papers that do not explicity mention network visualization. If it remains vague or unclear if network data is actually visualized} \highlight{INCLUDE} {\color{gray}the paper.\\
You MUST include papers describing e.g. a study, where a node-link diagram is viewed in virtual reality.}
\\\\
\highlight{
With your decision, you are the first filter step. All of your included papers will be checked again by another researcher.
Papers that are discarded by you will not be checked and are excluded.
Therefore, False-Negative classifications are much worse than False-Positives.
}
\\
{\color{gray}
\verb|[...]|
}
\\ \ 
\noindent\rule{\linewidth}{0.5pt}

\subsection*{\textbf{Prompt 3} ($P_3$), Changes to $P_{orig}$}

{
\color{gray}
\noindent
\verb|[...]|\\
Examples of immersive environments are:

\begin{itemize}
\item virtual reality (VR)
\item augmented reality (AR)
\item mixed reality (MR)
\item extended reality (XR),
\item wall-sized displays, or CAVE setups
\end{itemize}
}
\noindent
\highlight{DISCARD papers if you are certain that they are not relevant for this topic and do not match the criteria.\\
Otherwise, INCLUDE the paper.}{\color{gray}
\\\\
Below is the title and abstract. You must only answer with INCLUDE or DISCARD and a 2 sentence reason of why.
}
\\ \ 
\noindent\rule{\linewidth}{0.5pt} 

\subsection*{\textbf{Prompt 4} ($P_4$), Changes to $P_{orig}$}

{
\color{gray}
\noindent
\verb|[...]|\\
Examples of immersive environments are:

\begin{itemize}
\item virtual reality (VR)
\item augmented reality (AR)
\item mixed reality (MR)
\item extended reality (XR),
\item wall-sized displays, or CAVE setups
\end{itemize}
}
\noindent
\highlight{DISCARD papers if you are certain that they are not relevant for this topic and do not match the criteria.\\
Otherwise, INCLUDE the paper.
\\\\
With your decision, you are the first filter step. All of your included papers will be checked again by another researcher.\\
Papers that are discarded by you will not be checked and are excluded.\\
Therefore, False-Negative classifications are much worse than False-Positives.}{\color{gray}
\\\\
Below is the title and abstract. You must only answer with INCLUDE or DISCARD and a 2 sentence reason of why.
}
\\ \ 
\noindent\rule{\linewidth}{0.5pt} 

\subsection*{\textbf{Prompt 5} ($P_5$), Changes to $P_{orig}$}

{
\color{gray}
\noindent
\verb|[...]|\\
Examples of immersive environments are:

\begin{itemize}
\item virtual reality (VR)
\item augmented reality (AR)
\item mixed reality (MR)
\item extended reality (XR),
\item wall-sized displays, or CAVE setups
\end{itemize}
}
\noindent
\highlight{DISCARD papers if you are certain that they are not relevant for this topic and do not match the criteria.\\
Otherwise, INCLUDE the paper.
\\\\
With your decision, you are the first filter step. All of your included papers will be checked again by another researcher.\\
Papers that are discarded by you will not be checked and are excluded.\\
Therefore, False-Negative classifications are much worse than False-Positives.\\
However, keep in mind that the manual filtering effort should be kept as low as possible.\\
Thus, only include a paper if there is a chance that it might be of relevance.}{\color{gray}
\\\\
Below is the title and abstract. You must only answer with INCLUDE or DISCARD and a 2 sentence reason of why.
}
\\ \ 
\noindent\rule{\linewidth}{0.5pt} 

\subsection*{\textbf{Prompt 6} ($P_6$)}

You are a computer science professor conducting a literature review.
Your task is to decide and classify whether the given paper belongs to the research direction of ``visual network analysis in immersive environments.''
\\\\
You are provided with the title and abstract of a paper.
Your goal is to determine if the paper should be INCLUDED or DISCARDED.
\\\\
Decision Criteria:
You must INCLUDE a paper only if BOTH of the following are true:
\\\\
Network aspect is present: The paper explicitly deals with network data, graph data, or node-link diagrams (i.e., data visualizations representing relationships or connections). Examples include ``graph visualization'', ``network analysis'', ``node-link diagram'', ``social network graph'' etc.
\\\\
Immersive environment aspect is present: The paper explicitly involves immersive technologies or environments such as Virtual Reality (VR), Augmented Reality (AR), Mixed Reality (MR), Extended Reality (XR), or immersive visualization systems such as wall-sized displays, CAVEs, or head-mounted displays.
\\\\
Exclusion Rules:
You must DISCARD a paper if any of the following apply:
\\\\
It discusses only network visualization but not immersive environments.
\\\\
It discusses only immersive environments but not network or graph visualization.
\\\\
It refers to computer or communication networks (e.g., 5G, IoT, Wi-Fi, sensor, or data networks) instead of data visualization networks.
\\\\
It mentions ``network'' ambiguously (e.g., ``neural networks'', ``network traffic'', ``computational networks'', ``distributed networks'') without visualizing node-link data.
\\\\
It is vague or unclear whether network visualization is involved. When uncertain, DISCARD.
\\\\
Output Format:
Respond only with the following format:
INCLUDE or DISCARD
Reason: [exactly 1-2 sentences explaining your decision, referring to both aspects]
\\\\
Examples:
Example 1:
Title: ``Exploring Social Network Graphs in Virtual Reality''
- INCLUDE
Reason: The paper visualizes social network graphs (network data) using virtual reality (immersive environment).
\\\\
Example 2:
Title: ``Immersive Analytics of Scientific Data in AR''
- DISCARD
Reason: Although it uses AR, there is no mention of network or graph data visualization.
\\\\
Example 3:
Title: ``Optimizing 5G Network Topologies with AI''
- DISCARD
Reason: Refers to communication networks, not data visualization of node-link networks.
\\\\
When uncertain, err on the side of DISCARD.
\\ \ 
\noindent\rule{\linewidth}{0.5pt} 

\subsection*{\textbf{Prompt 7} ($P_7$), Changes to $P_6$}
{
\color{gray}
\noindent
\verb|[...]|\\
It is vague or unclear whether network visualization is involved. When uncertain, }\highlight{INCLUDE}.
\\
{
\color{gray}
\noindent
\verb|[...]|\\
When uncertain, err on the side of} \highlight{INCLUDE}.

\bibliographystyle{elsarticle-num} 
\bibliography{bibliography.bib}

\end{document}